%% file: neurips_2026.tex
\documentclass{article}

\PassOptionsToPackage{numbers,compress}{natbib}

 \usepackage[preprint]{neurips_2026}

\usepackage[utf8]{inputenc} % allow utf-8 input
\usepackage[T1]{fontenc}    % use 8-bit T1 fonts
\usepackage{hyperref}       % hyperlinks
\usepackage{url}            % simple URL typesetting
\usepackage{booktabs}       % professional-quality tables
\usepackage{amsfonts}       % blackboard math symbols
\usepackage{nicefrac}       % compact symbols for 1/2, etc.
\usepackage{microtype}      % microtypography
\usepackage{xcolor}         % colors
\usepackage{xspace}
\usepackage{multirow}
\usepackage{graphicx}
\usepackage{amsmath}
\usepackage[table]{xcolor}
\usepackage{graphicx}
\usepackage{makecell}
\usepackage{multirow}

\usepackage{algorithm}% http://ctan.org/pkg/algorithms
\usepackage{algpseudocode}% http://ctan.org/pkg/algorithmicx
\usepackage{xcolor}
\usepackage{fontawesome5}
\definecolor{darklinkcolor}{rgb}{0.10, 0.25, 0.55}

\title{Elephant in the Fridge: Constant-Memory Frame Packing for Long Video Understanding}

\newcommand{\model}{\texttt{FramePack-U}\xspace}

\author{%
  Shuo Gao \\
  \mbox{Renmin University of China}
  \And
  Chenhao Zheng \\
  \mbox{University of Washington} 
  \And
  Zhongzheng Ren \\
  \mbox{UNC-Chapel Hill} 
  \AND
  \vspace{-1.2em} \\
  \normalfont
  \makebox[\textwidth][c]{%
      \small
      \begin{tabular}{@{}c@{\hspace{0.45em}}l@{}}
        \raisebox{-0.1em}{\faHome} & \href{https://zeld0n.github.io/framepack-U/}{\texttt{\textcolor{darklinkcolor}{https://zeld0n.github.io/framepack-U/}}} \\[0.5em]
        \raisebox{-0.2em}{\includegraphics[height=1.05em]{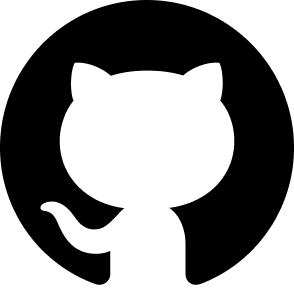}} & \href{https://github.com/ZelD0n/framepack-U}{\texttt{\textcolor{darklinkcolor}{https://github.com/ZelD0n/framepack-U}}} \\[0.5em]
        \raisebox{-0.4em}{\includegraphics[height=1.35em]{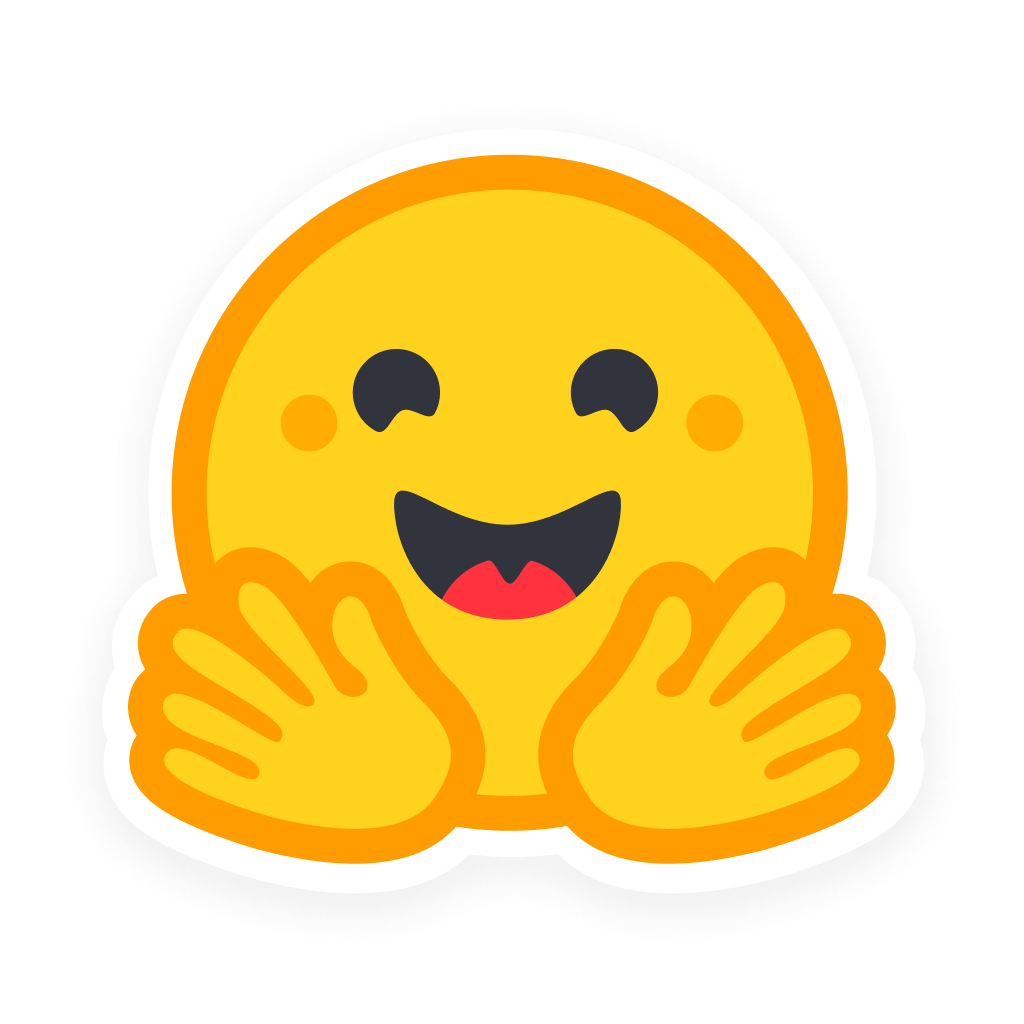}} & \href{https://huggingface.co/datasets/ZelD0n/FramePack-U}{\texttt{\textcolor{darklinkcolor}{https://huggingface.co/datasets/ZelD0n/FramePack-U}}} \\[0.4em]
        \raisebox{-0.25em}{\includegraphics[height=1.20em]{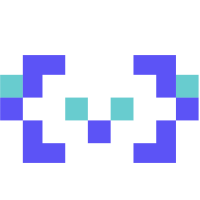}} & \href{https://www.modelscope.cn/datasets/ZelD0n/FramePack-U}{\texttt{\textcolor{darklinkcolor}{https://www.modelscope.cn/datasets/ZelD0n/FramePack-U}}}
      \end{tabular}%
    }
}

\begin{document}
\maketitle

\input{sections/abstract}
\input{sections/introduction}

\input{sections/relatedwork}

\input{sections/method}
\input{sections/experiments}
\input{sections/latency}

\input{sections/limitation}
\input{sections/conclusion}

\newpage
\bibliographystyle{plainnat}
\bibliography{references}

\newpage
\input{sections/appendix}

% \newpage
% \input{checklist.tex}

\end{document}

%% file: sections/abstract.tex
\begin{abstract}

Hour-long videos, streaming feeds, and movie-length content are pushing vision-language models (VLMs) past their breaking point—the standard recipe of encoding each frame into visual tokens yields sequences that grow linearly with video length, causing prohibitive memory consumption and inference latency. Existing approaches, such as token reduction and keyframe selection, attempt to shorten the token sequence but leave the $O(n)$ scaling intact, failing to resolve the fundamental bottleneck.
We propose \model, \emph{a strikingly simple, training-free frame packing framework that compresses arbitrarily long videos into a fixed-size visual representation at constant memory cost.} \model requires no fine-tuning or architectural modification — it plugs directly into off-the-shelf VLMs at inference time, yet consistently surpasses more elaborate baselines. 
Drawing inspiration from FramePack in generation literature, we propose a new content-aware compression strategy that operates under a fixed token budget for visual understanding.
Specifically, \model proceeds in two steps: (1) scoring each frame by its semantic similarity to the text prompt and ranking frames in a diversity-driven manner, and (2) packing the selected frames at varying resolutions into a fixed-length context window. This design concentrates the memory budget on frames that are both instruction-relevant and visually diverse.
While in theory \model can pack arbitrarily many frames into the fixed budget, we identify an empirical instantiation that generalizes robustly across different settings.
Extensive experiments on LongVideoBench, MLVU, and Video-MME show that \model consistently outperforms full-resolution baselines as well as prior token reduction and keyframe selection methods, across diverse model families, parameter scales, and frame budgets. Notably, on Qwen3.5-35B-A3B with a 128-frame budget, \model yields gains of \textbf{+12.9\%, +10.6\%, and +9.7\%} over the official model on the three benchmarks, respectively.

\end{abstract}

%% file: sections/introduction.tex
\section{Introduction}
\label{sec:intro}

Long video understanding underpins a wide range of real-world applications, including surveillance analysis, autonomous driving, sports commentary, movie comprehension, and instructional or educational assistants. Many of these scenarios further demand streaming settings, where models must continuously ingest incoming frames and respond on the fly rather than process a bounded clip offline. Unlike images or short videos, hour-long or unbounded streams carry rich temporal dynamics, evolving narratives, and long-horizon causal dependencies, all of which are essential for higher-level reasoning. Recent vision-language models (VLMs)~\citep{bai2025qwen3vltechnicalreport,wang2025internvl3,gemini3,openai2023gpt4,clark2025molmo2_cvpr26} offer a promising path forward by encoding visual content into tokens that are jointly processed with text instructions alongside an LLM backbone. Yet when scaled from images and short clips to long or streaming videos, this paradigm encounters a fundamental obstacle: the number of visual tokens grows proportionally with the number of frames, which severely dilutes context, inflates memory consumption, and drives inference cost up linearly or higher with video duration.

To address this, a large body of work resorts to information compression, typically through token reduction~\citep{fastvid, visionzip, framefusion} or keyframe selection~\citep{aks,videotree,focus}. By scoring visual features against handcrafted or learned criteria, these methods filter out non-essential content and retain only the clues deemed necessary for understanding. The VLM then operates on a much smaller set of visual tokens, yielding faster inference and reduced memory footprint without retraining the backbone.
However, such compression schemes share two fundamental limitations. (1) They only postpone the scalability bottleneck rather than eliminate it: the retained token count still scales as $O(n)$ or higher with video duration, so memory and latency continue to grow without bound as videos get longer. (2) Hard-dropping tokens or frames is a discrete and inherently lossy operation. Forcing a binary keep-or-discard decision fragments temporal continuity and erases transient actions that are often pivotal for long-horizon reasoning -- precisely the regime where long-video understanding matters most.

In this work we propose \model, \emph{a strikingly simple, training-free frame compression and packing method that extracts a fixed-size representation from videos of arbitrary length, achieving constant $O(1)$ memory complexity.} \model requires no fine-tuning or architectural modification—it plugs directly into off-the-shelf VLMs at inference time.
The key idea is to assign heterogeneous spatial resolutions to frames under a fixed total token budget regardless of video duration, so that the memory budget is spent on frames that matter most.
Inspired by FramePack \citep{framepack}, which packs frames at varying resolutions for efficient video generation, we adapt this idea for long video understanding. However, the original formulation is designed for generation, with a schedule favoring temporal continuity and proximity, hence unsuitable for understanding tasks that must preserve task-relevant evidence throughout the video.

To mitigate this issue, \model is designed to be content-aware, and operates in two steps.
(1) Relevance- and diversity-aware sampling. We compute the semantic similarity between the text prompt and each frame, producing a relevance score that identifies frames containing instruction-relevant evidence. Starting from the highest-scoring frame as a seed, we iteratively rank the remaining frames by selecting, at each step, the frame most visually distinct from those already chosen. This produces a single ordering that balances instruction relevance with visual diversity (Sec.~\ref{subsec:frame_sampling}). 
(2) Hierarchical resolution packing. We then assign frames along this ranking to resolution hierarchies, with top-ranked frames preserved at finer spatial resolutions and lower-ranked frames progressively downsampled (Sec.~\ref{subsec:hierarchical_pack}).
In principle, all frames can be retained, since the tiered scheme packs them into a fixed-size context regardless of video length—concentrating the memory budget on diverse, instruction-relevant frames while preserving full temporal coverage. While many tiered packing schedules are possible in principle, we provide an empirical instantiation, identified through experiments, that generalizes robustly across architectures, scales, and benchmarks.

We conduct extensive experiments on major long video VQA benchmarks such as  LongVideoBench~\citep{longvideobench}, MLVU~\citep{mlvu}, and Video-MME~\citep{videomme}. Our results have verified the effectiveness of \model against a number of competitive baselines. Under the same inference memory budgets, \model consistently outperforms full-resolution baselines across a variety of model series, parameter scales, and frame number settings. 
For instance, on the Mixture-of-Expert (MoE) model \texttt{Qwen3.5-35B-A3B} with a 128-frame budget, \model yields significant gains of \textbf{12.9\%, 10.6\%, and 9.7\%} on LongVideoBench, MLVU, and Video-MME (long), respectively. On an alternative backbone \texttt{Qwen3.6-27B}, \model yields consistent gains of \textbf{6.4\%, 8.0\%, and 2.7\%}.
Also, it surpasses existing token reduction and keyframe selection methods, demonstrating a better accuracy--efficiency trade-off in long and extreme-long video settings. It is worth noting that \model is training-free and directly applicable to off-the-shelf VLMs.

%% file: sections/relatedwork.tex
\section{Related Work}
\label{sec:relatedwork}

\noindent\textbf{Vision-Language Models.} 
VLMs have emerged in recent years as the dominant approach for image understanding and grounding~\citep{molmov1,openai2023gpt4,team2025gemma,gemini3}. The prevailing recipe pairs a CLIP-style~\citep{clip,siglip2} visual encoder with an LLM backbone, bridging the two through a lightweight connector that projects image embeddings into the LLM token space~\citep{molmov1,liu2024llava665k}. Video VLMs~\citep{eagle2_5, clark2025molmo2_cvpr26, bai2025qwen3vltechnicalreport, cho2025PerceptionLM, maaz2024videochatgpt} typically inherit this design and apply the image encoder frame-by-frame to obtain per-frame embeddings. A handful of efforts replace or augment the per-frame pipeline with pretrained video encoders, often operating on frame pairs or short clips~\citep{apollo, wang2024qwen2vl, glmv}, but scaling video encoders to longer temporal windows still trails per-frame image encoding with strong backbones such as SigLIP 2~\citep{siglip2}.

\noindent\textbf{Long Video VLM.}
To curb the token overhead of frame-wise encoding, a growing body of work pursues more compact video representations. One line of research performs \emph{token reduction} inside 
the visual stream, merging or pruning redundant tokens based on attention scores, similarity, or learned importance~\citep{shen2024longvu, li2024videochat, wang2025internvl3}. The second line focuses on \emph{keyframe selection}, scoring frames by query relevance or diversity and discarding the rest before they reach the LLM~\citep{aks, 
videotree, focus, yang2025kwai}. Other works explore architectural changes, including memory banks for streaming video~\citep{zhang2024flashvstream}, sequential temporal aggregation~\citep{weng2024longvlm}, and slow--fast pathways that mix high-fidelity keyframes with coarsely encoded context frames~\citep{xu2024slowfast, apollo}. Despite their differences, these approaches share a common property: the resulting token sequence still scales proportionally with video duration, so memory and latency continue to grow as videos get longer. 

\noindent\textbf{Packing for Generation.}
FramePack~\citep{framepack} proposes a context packing strategy for next-frame prediction models in video generation. The key idea is to assign heterogeneous token budgets to input frames according to their importance, with frames closer to the prediction target receiving longer contexts via finer-grained patchifying kernels. As a result, the generation workload becomes invariant to video length, enabling long video synthesis on consumer GPUs. The packing principle has since been adopted beyond video generation: FastAvatar~\citep{fastavatar} applies it to 3D avatar reconstruction, retaining a sparse set of high-fidelity keyframes whilst compressing the remaining frames into an aggregated token representation, enabling feedforward reconstruction from hundreds of frames within a fixed memory footprint. Yume~\citep{yume} and Yume1.5~\citep{yume15} adopt related history-compression strategies for interactive world generation, where historical frames are compressed at different rates to support long-form autoregressive generation. \model shares the same packing philosophy, i.e., allocating more tokens to more important frames within a fixed budget. However, rather than generation, it targets \emph{understanding}, adapting the packing strategy to the inference-time encoding of video inputs in VLMs.

%% file: sections/method.tex
% \begin{figure}[t]
% \centering
% \includegraphics[width=1.0\linewidth]{figures/framepack_framework.pdf}
% \caption{\textbf{Overview of \model.}
% Given a long video and a text query, \model selects query-relevant and visually diverse frames, then packs them with heterogeneous resolutions under a fixed memory budget. Higher-priority frames are kept at higher resolution, while lower-priority frames are progressively compressed. The packed frames are fed into the VLM for answer generation.}
% \label{fig:framework}
% \end{figure}
% \begin{figure}[t]
% \centering
% \includegraphics[width=0.8\linewidth]{figures/schemes.pdf}
% \caption{\textbf{Illustration of packing schemes under the same memory budget.}
% Using a 4-frame memory budget as an example, red, yellow, and blue denote frames with downsampling factors $d=1,2,4$, respectively. Different resolution combinations can use the same memory while packing different numbers of frames.}
% \label{fig:packing_scheme}
% \end{figure}
% \section{\model}
% \label{sec:method}
\begin{figure}[t]
  \centering
  % --- 第一张图 (Figure 1) ---
  \includegraphics[width=0.98\linewidth]{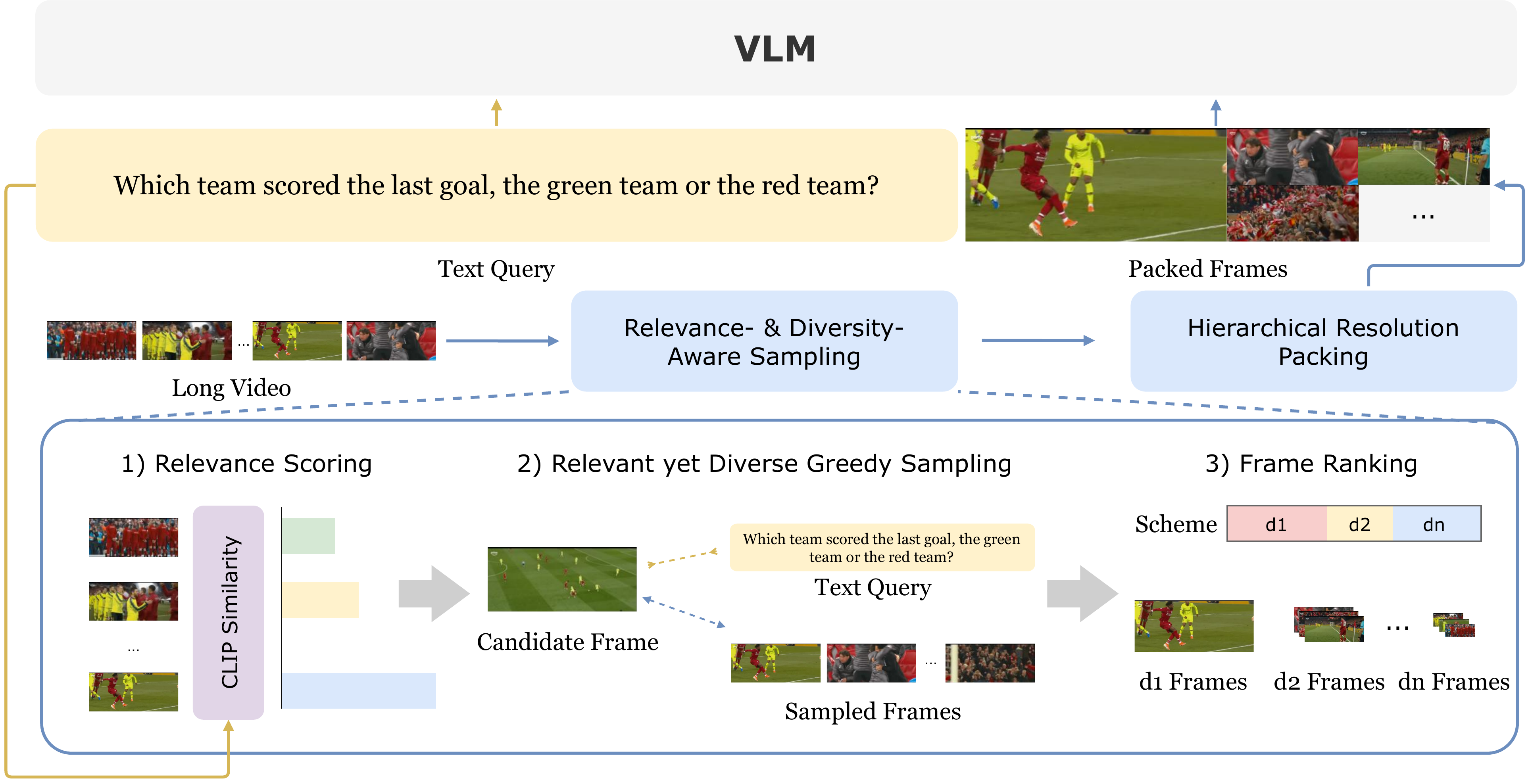}
  % \vspace{-0.4em}
  \caption{\textbf{Overview of \model.}
  Given a long video and a text query, \model selects query-relevant and visually diverse frames, then packs them with heterogeneous resolutions under a fixed memory budget. Higher-priority frames are kept at higher resolution, while lower-priority frames are progressively compressed. The packed frames are fed into the VLM for answer generation.}
  \label{fig:framework}

  \vspace{2.5em} % 控制两张图及图注之间的垂直间隙

  % --- 第二张图 (Figure 2) ---
  \includegraphics[width=0.78\linewidth]{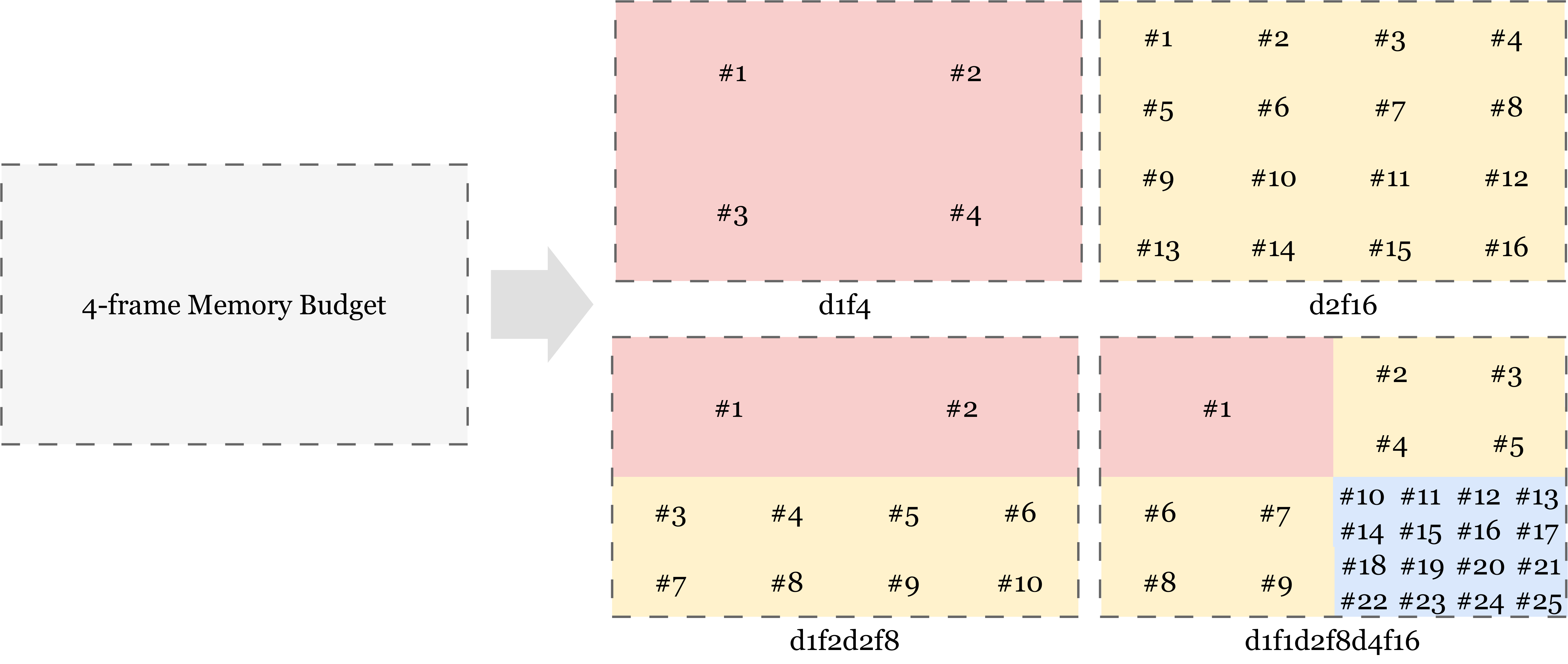}
  % \vspace{-0.4em}
  \caption{\textbf{Illustration of packing schemes under the same memory budget.}
  Using a 4-frame memory budget as an example, red, yellow, and blue denote frames with downsampling factors $d=1,2,4$, respectively. Different resolution combinations can use the same memory while packing different numbers of frames.}
  \label{fig:packing_scheme}
\end{figure}

\section{\model}
\label{sec:method}

Let $\mathcal{V} = \{\mathbf{x}_t\}_{t=1}^{T}$ denote a video of $T$ frames. Our goal is to construct a packed visual sequence $\mathcal{V}^* = \{\mathbf{x}^*_t\}_{t=1}^{N}$ whose total visual token count is bounded by a fixed memory budget, independent of the video length $T$. 
As shown in Fig~\ref{fig:framework}, \model proceeds in 2 stages: (1) \textbf{relevance- and diversity-aware sampling}, which reduces $\mathcal{V}$ to $\mathcal{V}^*$ candidate frames via retrieval guided by instruction relevance and visual diversity (Sec.~\ref{subsec:frame_sampling}); and (2) \textbf{hierarchical resolution packing}, which assigns each selected frame a spatial downsampling factor according to its estimated priority, subject to the budget constraint (Sec.~\ref{subsec:hierarchical_pack}). The packed sequence is then passed to the VLM backbone as visual tokens.

\noindent\textbf{Discussion of the necessity of $N$.} In principle, $N$ can be as large as $T$, so \model can pack a video of arbitrary length into the fixed budget. In practice, we treat $N$ as a tunable parameter that controls how many candidate frames are considered for packing, giving the framework flexibility and enabling systematic study of how input scale affects performance.
Later in experiments we show that performance is not monotonically increasing in $N$. Under a fixed token budget, larger $N$ improves temporal coverage but forces lower per-frame resolution, and most VLMs degrade once frame resolution falls below the range seen during pre-training. The optimal $N$ thus reflects a trade-off between temporal density and per-frame fidelity, and depends on model capacity and pre-training data; we report the setting that generalizes best across our evaluations in Sec.~\ref{sec:experiments}.

\subsection{Relevance- and diversity-aware frame sampling}
\label{subsec:frame_sampling}

We adopt a relevance and diversity-aware selection strategy over these $N$ candidates that favors frames that are both semantically relevant to the text instruction and visually non-redundant. 
Concretely, we first use CLIP~\citep{clip} to compute the cosine similarity $r_i=\operatorname{CLIP}(\mathbf{x}_i, \mathbf{q})$ between  frame $\mathbf{x}_i$ and the input text prompt $\mathbf{q}$.

We then start with the greedy sampling process. Let $n\in N$ denote the current sampling step, and $\tilde{\mathcal{V}}_{n}$ denote the set of chosen frames before the $n$-th step. 
The next frame is sampled as:
\begin{equation}
\mathbf{x}_{n}^{*} =
\begin{cases}
\displaystyle
\arg\max_{\mathbf{x}_i \in \mathcal{V}} r_i, & n = 1, \\[5pt]
\displaystyle
\arg\max_{\mathbf{x}_i \in \mathcal{V} \setminus \tilde{\mathcal{V}}}
\left[ \lambda \cdot r_i - (1-\lambda) \max_{\mathbf{x}_j \in \tilde{\mathcal{V}}} \operatorname{sim}(\mathbf{x}_i, \mathbf{x}_j) \right], & n\in [2,N]
\end{cases}
\label{eq:diversity}
\end{equation}
After each step, we update $\tilde{\mathcal{V}}=\tilde{\mathcal{V}}\cup\{\mathbf{x}_{n}^{*}\}$ until $N$ frames are selected. This greedy procedure ensures that the retained frames cover the most instruction-relevant content while avoiding the over-selection of visually redundant frames.

\subsection{Hierarchical resolution packing}
\label{subsec:hierarchical_pack}

Given the $N$ ranked candidate frames from Sec.~\ref{subsec:frame_sampling}, \model assigns each one a spatial resolution under a fixed token budget. The guiding principle is that frames differ in their relevance to the query and their representativeness of the video: rather than encoding all frames at uniform fidelity, \model allocates higher spatial resolution to more informative frames and progressively compresses the remainder, maximizing temporal coverage within the budget. We next describe the packing scheme, the frame ranking it relies on, and the concrete resolution schedule we adopt.

\paragraph{Packing scheme.} A packing scheme is a hierarchical resolution assignment
\begin{equation}
    \mathcal{S} = \{(d,\, f_d)\}_{d \in \{1, 2, 4, 8, \ldots\}},
\end{equation}
where $d$ is a spatial downsampling factor and $f_d$ is the number of frames assigned to level $d$, with $\sum_{d} f_d = N$. Downsampling a frame by $d$ in each spatial dimension yields $M/d^2$ tokens per frame, where $M$ is the token count of a full-resolution frame. The total memory footprint of a scheme is therefore constrained by
\begin{equation}
     M \sum_{d} \frac{f_d}{d^2} \;\leq\; B,
    \label{eq:budget}
\end{equation}
where $B$ is the fixed token budget.

For convenience, we denote a level of $n$ frames at downsampling factor $k$ as \texttt{d\{$k$\}f\{$n$\}}, and concatenate segments to describe mixed-resolution schemes. For example, \texttt{d1f16d2f448} allocates $16$ frames at full resolution ($d{=}1$) and $448$ frames at $2{\times}$ downsampling ($d{=}2$), with a total cost of $16 + 448/4 = 128$ full-resolution frame equivalents.

\paragraph{Frame ranking.} The packing scheme determines \emph{how many} frames live at each resolution; a ranking determines \emph{which} frames they are. As illustrated in Figure~\ref{fig:packing_scheme}, different packing schemes can consume the same memory budget while retaining different numbers of frames by combining multiple resolution levels. We rank frames by the score in Eq.~\eqref{eq:diversity}: frames selected earlier in the greedy procedure receive higher priority and are assigned to finer-resolution levels (smaller $d$). This concentrates the high-resolution quota on frames that are both instruction-relevant and visually diverse. The full procedure is summarized in Alg.~\ref{alg:framepack}.

As a query-agnostic alternative, we also consider a binary-tree midpoint ordering that recursively ranks frames by temporal midpoints, ensuring uniform temporal coverage without relying on query signals. We include it as a baseline in our ablations (Sec.~\ref{sec:experiments}).

\begin{algorithm}[t]
\caption{\model}
\label{alg:framepack}
\begin{algorithmic}[1]
\Require Video frames $\mathcal{V}=\{x_i\}_{i=1}^{T}$, query $q$, packed frame number $N$, packing scheme $\mathcal{S}=\{(d,f_d)\}$, trade-off coefficient $\lambda$
\Ensure Packed visual sequence $\mathcal{V}^{*}$

\State Compute relevance scores $r_i \gets \mathrm{CLIP}(x_i,q)$ for all $x_i\in\mathcal{V}$
\State Initialize selected set $\widetilde{\mathcal{V}}\gets\emptyset$

\For{$n=1,\ldots,N$}
    \State Select
    \(
    x_n^{*} \gets
    \begin{cases}
    \arg\max_{x_i\in\mathcal{V}} r_i, & n=1,\\
    \arg\max_{x_i\in\mathcal{V}\setminus\widetilde{\mathcal{V}}}
    \left[
    \lambda r_i-(1-\lambda)\max_{x_j\in\widetilde{\mathcal{V}}}\mathrm{sim}(x_i,x_j)
    \right], & n>1.
    \end{cases}
    \)
    \State $\widetilde{\mathcal{V}}\gets\widetilde{\mathcal{V}}\cup\{x_n^{*}\}$
\EndFor

\State Sort packing levels in $\mathcal{S}$ by increasing downsampling factor $d$
\For{$n=1,\ldots,N$}
    \State Find $d$ such that $\sum_{d'<d} f_{d'} < n \leq \sum_{d'\leq d} f_{d'}$
    \State $\hat{x}_n \gets \mathrm{Resize}(x_n^{*},1/d)$
\EndFor

\State $\mathcal{V}^{*}\gets$ temporal-order sorting of $\{\hat{x}_n\}_{n=1}^{N}$
\State \Return $\mathcal{V}^{*}$
\end{algorithmic}
\end{algorithm}

%% file: sections/experiments.tex
\section{Experiments}
\label{sec:experiments}
In this section, we evaluate the effectiveness of \model on long video understanding tasks. We first describe the experimental setup. We then compare \model with existing training-free video compression methods (Sec.~\ref{subsec:results}) . Finally, we conduct ablation studies (Sec.~\ref{subsec:ablation}) and show qualitative results (Sec.~\ref{subsec:qualitative}).

\noindent\textbf{Benchmarks.}
We evaluate our method on three most popular long video understanding benchmarks:
LongVideoBench~\citep{longvideobench}, MLVU~\cite{mlvu}, and Video-MME~\cite{videomme}. LongVideoBench reports the overall accuracy as well as results under different video-duration ranges, i.e., 15s, 60s, 600s, and 3600s. MLVU reports the M-average score. For Video-MME, we report the overall accuracy and the performance on short, medium, and long videos, with particular emphasis on the long-video split.

\noindent\textbf{Base models.}
To validate the effectiveness of our method across different scales and architectures, we conduct experiments on two Qwen VLMs: \texttt{Qwen3.6-27B} (dense) and \texttt{Qwen3.5-35B-A3B} (MoE). 

\noindent\textbf{Compared methods.}
We first compare against the vanilla uniform sampling strategy suggested by the official guidance. 
Additionally, we compare to representative training-free video token/frame reduction methods, including FastVID~\citep{fastvid}, VisionZip~\citep{visionzip}, and FrameFusion~\citep{framefusion}. These methods reduce the visual computation by selecting or compressing visual tokens/frames before feeding them into the language model.

\noindent\textbf{Implementation details.}
Given an input video, we uniformly sample $T$ frames as the initial frame set. Our method then assigns each frame to different resolution levels under a fixed memory budget. Unless otherwise specified, we use the diversity-aware packing strategy as our default method. The strategy first measures query-frame relevance using CLIP similarity and then performs diversity-aware selection based on frame-level visual feature similarity, preventing visually redundant frames from occupying the high-resolution budget. For the CLIP model, we choose CLIP-ViT-B/32 for its lightweight nature. $\lambda$ is set to 0.7. For the initial sampling rate, we sample all frames in an offline way for the benchmarking experiments. It could also be set in online mode for lower latency. We use the global feature extracted by ViT for frame-to-frame similarity. Regarding the resized dimensions, we maintain the original resolutions following the default settings for LongVideoBench and Video-MME. For MLVU, we proportionally resize candidate frames such that the longer edge is 1280 pixels before subsequent processing. We follow the practice of the benchmarks and directly use their testing text prompts.

\subsection{Results}
\label{subsec:results}
We compare \model with representative training-free video compression methods under the same visual memory budgets. Specifically, we evaluate all methods using 128-, 256-, and 512-frame memory, where each budget denotes the memory cost equivalent to processing the corresponding number of full-resolution frames. This setting allows us to assess whether different methods can improve video understanding accuracy without increasing the visual memory cost.

\begin{table*}[t]
\centering
\begin{minipage}[t]{0.49\textwidth}
\centering
% \tiny
\scriptsize
\setlength{\tabcolsep}{3pt}
\begin{tabular}{llll}
% \toprule
\textbf{Method} & \textbf{LVB} & \textbf{MLVU} & \textbf{V-MME} \\
\midrule
\multicolumn{4}{c}{Proprietary models} \\
\midrule
\rowcolor{gray!15} GPT-5~\citep{gpt5} & 72.6 & 77.7 & 83.3 \\
\rowcolor{gray!15} GPT-5 mini~\citep{gpt5} & 69.7 & 69.1 & 77.3 \\
\rowcolor{gray!15} Gemini 3 Pro~\citep{gemini3} & 75.9 & 75.7 & 88.6 \\
\rowcolor{gray!15} Gemini 2.5 Flash~\citep{gemini25} & 73.1 & 75.1 & 84.2 \\
\rowcolor{gray!15} Claude Sonnet 4.5~\citep{anthropic2025sonnet} & 65.1 & 64.0 & 74.2 \\
% \rowcolor{gray!15} InternVL3.5-241B-A28B & 67.1 & 78.2 & 72.9 \\
\midrule
\multicolumn{4}{c}{128 frames} \\
\midrule
\rowcolor{gray!15}
Vanilla & 68.0 & 75.9 & 73.6 \\
FastVID~\citep{fastvid} & 59.9{\color{red}{\tiny$_{-8.1}$}} & 67.3{\color{red}{\tiny$_{-8.6}$}} & 61.0{\color{red}{\tiny$_{-12.6}$}} \\
VisionZip~\citep{visionzip} & 57.7{\color{red}{\tiny$_{-10.3}$}} & 67.4{\color{red}{\tiny$_{-8.5}$}} & 63.2{\color{red}{\tiny$_{-10.4}$}} \\
FrameFusion~\citep{framefusion} & 57.8{\color{red}{\tiny$_{-10.2}$}} & 65.4{\color{red}{\tiny$_{-10.5}$}} & 63.5{\color{red}{\tiny$_{-10.1}$}} \\
\model(Ours) & \underline{74.4}{\color{green!50!black}{\tiny$_{+6.4}$}} & \textbf{83.9}{\color{green!50!black}{\tiny$_{+8.0}$}} & \underline{76.3}{\color{green!50!black}{\tiny$_{+2.7}$}} \\
\midrule
\multicolumn{4}{c}{256 frames} \\
\midrule
\rowcolor{gray!15}
Vanilla & 70.9 & 76.7 & 75.9 \\
FastVID~\citep{fastvid} & 60.8{\color{red}{\tiny$_{-10.1}$}} & 68.9{\color{red}{\tiny$_{-7.8}$}} & 62.2{\color{red}{\tiny$_{-13.7}$}} \\
VisionZip~\citep{visionzip} & 62.8{\color{red}{\tiny$_{-8.1}$}} & 70.7{\color{red}{\tiny$_{-6.0}$}} & 65.2{\color{red}{\tiny$_{-10.7}$}} \\
FrameFusion~\citep{framefusion} & 61.6{\color{red}{\tiny$_{-9.3}$}} & 67.7{\color{red}{\tiny$_{-9.0}$}} & 65.4{\color{red}{\tiny$_{-10.9}$}} \\
\model(Ours) & 74.3{\color{green!50!black}{\tiny$_{+3.4}$}} & \underline{83.2}{\color{green!50!black}{\tiny$_{+6.5}$}} & \textbf{77.9}{\color{green!50!black}{\tiny$_{+2.0}$}} \\
\midrule
\multicolumn{4}{c}{512 frames} \\
\midrule
\rowcolor{gray!15}
Vanilla & 71.8 & 78.0 & 74.4 \\
FastVID~\citep{fastvid} & 58.5{\color{red}{\tiny$_{-13.3}$}} & 68.4{\color{red}{\tiny$_{-9.6}$}} & 61.1{\color{red}{\tiny$_{-13.3}$}} \\
VisionZip~\citep{visionzip} & 63.7{\color{red}{\tiny$_{-8.1}$}} & 69.9{\color{red}{\tiny$_{-8.1}$}} & 67.2{\color{red}{\tiny$_{-7.2}$}} \\
FrameFusion~\citep{framefusion} & 62.8{\color{red}{\tiny$_{-9.0}$}} & 70.5{\color{red}{\tiny$_{-7.5}$}} & 66.9{\color{red}{\tiny$_{-7.5}$}} \\
\model(Ours) & \textbf{74.9}{\color{green!50!black}{\tiny$_{+3.1}$}} & 82.1{\color{green!50!black}{\tiny$_{+4.1}$}} & \underline{76.3}{\color{green!50!black}{\tiny$_{+1.9}$}} \\
% \bottomrule
\end{tabular}
\end{minipage}
\hfill
\begin{minipage}[t]{0.49\textwidth}
\centering
% \tiny
\scriptsize
\setlength{\tabcolsep}{3pt}
\begin{tabular}{llll}
% \toprule
\textbf{Method} & \textbf{LVB} & \textbf{MLVU} & \textbf{V-MME} \\
\midrule
\multicolumn{4}{c}{Proprietary models} \\
\midrule
\rowcolor{gray!15} GPT-5~\citep{gpt5} & 72.6 & 77.7 & 83.3 \\
\rowcolor{gray!15} GPT-5 mini~\citep{gpt5} & 69.7 & 69.1 & 77.3 \\
\rowcolor{gray!15} Gemini 3 Pro~\citep{gemini3} & 75.9 & 75.7 & 88.6 \\
\rowcolor{gray!15} Gemini 2.5 Flash~\citep{gemini25} & 73.1 & 75.1 & 84.2 \\
\rowcolor{gray!15} Claude Sonnet 4.5~\citep{anthropic2025sonnet} & 65.1 & 64.0 & 74.2 \\
% \rowcolor{gray!15} InternVL3.5-241B-A28B & 67.1 & 78.2 & 72.9 \\
\midrule
\multicolumn{4}{c}{128 frames} \\
\midrule
\rowcolor{gray!15}
Vanilla & 59.4 & 71.8 & 66.7 \\
FastVID~\citep{fastvid} & 50.3{\color{red}{\tiny$_{-9.1}$}} & 66.7{\color{red}{\tiny$_{-5.1}$}} & 61.5{\color{red}{\tiny$_{-5.2}$}} \\
VisionZip~\citep{visionzip} & 52.2{\color{red}{\tiny$_{-7.2}$}} & 65.6{\color{red}{\tiny$_{-6.2}$}} & 60.7{\color{red}{\tiny$_{-6.0}$}} \\
FrameFusion~\citep{framefusion} & 56.5{\color{red}{\tiny$_{-2.9}$}} & 63.9{\color{red}{\tiny$_{-7.9}$}} & 61.3{\color{red}{\tiny$_{-5.4}$}} \\
\model(Ours) & 72.3{\color{green!50!black}{\tiny$_{+12.9}$}} & 82.4{\color{green!50!black}{\tiny$_{+10.6}$}} & \textbf{76.4}{\color{green!50!black}{\tiny$_{+9.7}$}} \\
\midrule
\multicolumn{4}{c}{256 frames} \\
\midrule
\rowcolor{gray!15}
Vanilla & 60.3 & 73.0 & 67.1 \\
FastVID~\citep{fastvid} & 51.2{\color{red}{\tiny$_{-9.1}$}} & 68.4{\color{red}{\tiny$_{-4.6}$}} & 63.9{\color{red}{\tiny$_{-3.2}$}} \\
VisionZip~\citep{visionzip} & 56.9{\color{red}{\tiny$_{-3.4}$}} & 67.5{\color{red}{\tiny$_{-5.5}$}} & 64.6{\color{red}{\tiny$_{-2.5}$}} \\
FrameFusion~\citep{framefusion} & 57.5{\color{red}{\tiny$_{-2.8}$}} & 67.7{\color{red}{\tiny$_{-5.3}$}} & 64.8{\color{red}{\tiny$_{-2.3}$}} \\
\model(Ours) & \underline{72.6}{\color{green!50!black}{\tiny$_{+12.3}$}} & \underline{82.9}{\color{green!50!black}{\tiny$_{+9.9}$}} & 75.5{\color{green!50!black}{\tiny$_{+8.4}$}} \\
\midrule
\multicolumn{4}{c}{512 frames} \\
\midrule
\rowcolor{gray!15}
Vanilla & 59.8 & 74.3 & 67.3 \\
FastVID~\citep{fastvid} & 57.9{\color{red}{\tiny$_{-1.9}$}} & 68.4{\color{red}{\tiny$_{-5.9}$}} & 64.5{\color{red}{\tiny$_{-2.8}$}} \\
VisionZip~\citep{visionzip} & 55.2{\color{red}{\tiny$_{-4.6}$}} & 69.4{\color{red}{\tiny$_{-4.9}$}} & 64.4{\color{red}{\tiny$_{-2.9}$}} \\
FrameFusion~\citep{framefusion} & 59.9{\color{green!50!black}{\tiny$_{+0.1}$}} & 69.3{\color{red}{\tiny$_{-5.0}$}} & 65.4{\color{red}{\tiny$_{-1.9}$}} \\
\model(Ours) & \textbf{73.0}{\color{green!50!black}{\tiny$_{+13.2}$}} & \textbf{84.7}{\color{green!50!black}{\tiny$_{+10.4}$}} & \underline{75.8}{\color{green!50!black}{\tiny$_{+8.5}$}} \\
% \bottomrule
\end{tabular}
\end{minipage}
\label{tab:results_main}
\caption{Main results on Qwen3.6-27B (\textbf{left}) and Qwen3.5-35B-A3B (\textbf{right}) across different memory budgets. Subscripts indicate the gap to Vanilla at the same budget; \textbf{bold} and \underline{underline} denote the best and second-best results among open-weight models, respectively. We find that: (1) \model outperforms all baselines substantially, (2) our top results are competitive with, and often surpasses, those of much larger proprietary VLMs.}
\end{table*}

\noindent\textbf{Results on Qwen3.6-27B.}
Table~\ref{tab:results_main} left compares our method with existing training-free compression baselines on Qwen3.6-27B. We use \texttt{d1f16d2f448}, \texttt{d1f250d4f96} and \texttt{d1f128d2f1536} for 128-, 256- and 512-frame memory, respectively.
Across three levels of memory usage, \model consistently achieves the best overall performance among all methods.
(1) Compared with the vanilla baseline, \model brings clear improvements across every memory budget, with the largest relative gains appearing under the tighter 128-frame memory setting.
(2) Compared with existing training-free compression methods, \model also shows stronger performance across benchmarks, demonstrating that preserving broad temporal coverage while adaptively allocating spatial resolution is more effective than directly dropping frames or compressing visual tokens.
These results suggest that \model makes better use of a fixed visual memory budget by selecting query-relevant and visually diverse frames for higher-resolution processing.

\noindent\textbf{Results on Qwen3.5-35B-A3B.}
Table~\ref{tab:results_main} right reports the results on Qwen3.5-35B-A3B under different memory budgets. We use \texttt{d1f32d2f384}, \texttt{d1f64d2f768} and \texttt{d1f384d2f512} for 128-, 256- and 512-frame memory, respectively.
Similarly, \model consistently outperforms the vanilla baseline and achieves the best overall performance among all compared methods.
The improvements are particularly notable because Qwen3.5-35B-A3B has a weaker vanilla baseline than Qwen3.6-27B, yet \model substantially narrows the performance gap by better utilizing the fixed visual memory budget.
These results further demonstrate the robustness of \model across different architectures and confirm the benefit of prioritizing query-relevant and visually diverse frames.

\subsection{Ablations}
\label{subsec:ablation}

\paragraph{Ablation study on schemes.}
The hierarchical packing scheme determines the trade-off between temporal coverage and per-frame spatial fidelity.
We conduct a broad search over mixed-resolution schemes and identify several configurations that work consistently well across benchmarks.
Table~\ref{tab:ablation_schemes} reports representative strong schemes under the 512-frame memory budget on Qwen3.6-27B.

Overall, \model is robust to the exact choice of scheme.
All mixed-resolution schemes in Table~\ref{tab:ablation_schemes} clearly outperform the vanilla baseline under the same memory budget.
Different schemes slightly favor different benchmarks: for example, \texttt{d1f128d2f1536} performs best on LongVideoBench, \texttt{d1f384d2f512} performs best on MLVU, and \texttt{d1f480d2f128} performs best on Video-MME.
This suggests that there is no single universally optimal allocation, as different tasks may require different balances between fine-grained spatial evidence and broader temporal coverage.

Based on this search, we select schemes that achieve strong overall performance while keeping a balanced allocation between high-resolution anchor frames and low-resolution contextual frames.
Importantly, the gains are not tied to a single carefully tuned configuration: even when the number of packed frames varies substantially, from 608 to 1664 under the 512-frame budget, \model maintains consistent improvements over the vanilla baseline.
This robustness supports our design choice of hierarchical resolution packing and shows that the key benefit comes from preserving broad temporal context while assigning higher fidelity to informative frames.

\begin{table*}[t]
\centering
\begin{minipage}[t]{0.49\textwidth}
\vspace{0pt}
\centering
\scriptsize
\setlength{\tabcolsep}{3pt}
\begin{tabular}{llcccc}
% \toprule
\textbf{Memory} & \textbf{Scheme} & \textbf{\#Frame} & \textbf{LVB} & \textbf{MLVU} & \textbf{V-MME} \\
\midrule
\rowcolor{gray!15}
\multirow{5}{*}{512 frames}
& Vanilla & 512 & 71.8 & 78.0 & 74.4 \\
& \texttt{d1f128d2f1536} & 1664 & \textbf{74.9} & 82.1 & 76.3 \\
& \texttt{d1f256d2f1024} & 1280 & 74.7 & 82.2 & 75.9 \\
& \texttt{d1f384d2f512} & 896 & 74.6 & \textbf{82.9} & 76.0 \\
& \texttt{d1f480d2f128} & 608 & 74.3 & 81.0 & \textbf{76.5} \\
% \bottomrule
\end{tabular}
\caption{\textbf{Ablation on different resolution allocation schemes using Qwen3.6-27B}. See Appendix~\ref{app:more_ablation} for ablations on 128- and 256-frame budgets.}
\label{tab:ablation_schemes}
\end{minipage}
\hfill
\begin{minipage}[t]{0.49\textwidth}
\vspace{0pt}
\centering
\scriptsize
\setlength{\tabcolsep}{4pt}
\begin{tabular}{clccc}
% \toprule
\textbf{Memory} & \textbf{Model} & \textbf{LVB} & \textbf{MLVU} & \textbf{V-MME} \\
\midrule
\multirow{3}{*}{\makecell{512 frames \\ (\texttt{d1f256d2f1024})}}
& Qwen3.5-4B & +0.9 & -1.0 & -1.2 \\
& Qwen3.5-9B & +1.6 & -0.6 & -0.4 \\
& Qwen3.6-27B & \textbf{+2.5} & \textbf{+4.2} & \textbf{+1.5} \\
% \bottomrule
\end{tabular}
\caption{\textbf{Ablation on model size.} We report the performance gain of \model over the vanilla baseline across different model sizes. 
See Appendix~\ref{app:more_ablation} for ablations on 128- and 256-frame budgets.}
\label{tab:ablation_model_size}
\end{minipage}
\end{table*}

\paragraph{Ablation study on frame diversity.}

\begin{table}[t]
\centering
\scriptsize
\setlength{\tabcolsep}{3pt}
\begin{tabular}{llllll}
% \toprule
\textbf{Memory} & \textbf{Scheme} & \textbf{Strategy} & \textbf{LongVideoBench} & \textbf{MLVU} & \textbf{Video-MME} \\
\midrule
\rowcolor{gray!15}
\multirow{10}{*}{512 frames}
& -- & vanilla & 71.8 & 78.0 & 74.4 \\
& \multirow{3}{*}{\texttt{d1f128d2f1536}}
& \model & \textbf{74.9} & 82.1 & \textbf{76.3} \\
& & \quad$-$ Relevance & 74.0 {\color{red}(-0.9)} & 82.3 {\color{green!50!black}(+0.2)} & \textbf{76.3} {\color{gray}(+0.0)} \\
& & \quad$-$ Diversity & 73.9 {\color{red}(-1.0)} & 80.4 {\color{red}(-1.7)} & 69.8 {\color{red}(-6.5)} \\
\cmidrule(lr){3-6}
& \multirow{3}{*}{\texttt{d1f256d2f1024}}
& \model & 74.7 & 82.2 & 75.9 \\
& & \quad$-$ Relevance & 74.6 {\color{red}(-0.1)} & 82.3 {\color{green!50!black}(+0.1)} & 76.0 {\color{green!50!black}(+0.1)} \\
& & \quad$-$ Diversity & 73.8 {\color{red}(-0.9)} & 80.6 {\color{red}(-1.6)} & 69.6 {\color{red}(-6.3)} \\
\cmidrule(lr){3-6}
& \multirow{3}{*}{\texttt{d1f384d2f512}}
& \model & 74.6 & \textbf{82.9} & 76.0 \\
& & \quad$-$ Relevance & 73.1 {\color{red}(-1.5)} & 81.1 {\color{red}(-1.8)} & 75.4 {\color{red}(-0.6)} \\
& & \quad$-$ Diversity & 73.7 {\color{red}(-0.9)} & 80.8 {\color{red}(-2.1)} & 69.9 {\color{red}(-6.1)} \\
% \bottomrule
\end{tabular}
\caption{\textbf{Ablation on prompt relevance and frame diversity using Qwen3.6-27B}. \textit{$-$ Relevance} removes prompt relevance and uses a prompt-agnostic diversity-based ordering, while \textit{$-$ Diversity} ranks frames only by query-frame similarity. \model combines prompt relevance and visual diversity. Values in parentheses denote the difference between each ablated variant and \model. See Appendix~\ref{app:more_ablation} for ablations on 128- and 256-frame budgets.}
\label{tab:ablation_relevance_diversity}
\end{table}

We further study the effect of the diversity-aware frame selection strategy on Qwen3.6-27B.
As an alternative, we consider a \emph{Relevance-only} strategy, which ranks frames solely according to their semantic similarity with the text query and assigns higher resolutions to the top-ranked frames.
While this strategy can identify query-relevant frames, it does not explicitly consider visual redundancy among selected frames.
As a result, multiple high-resolution slots may be allocated to visually similar frames, limiting the effective temporal coverage.

Table~\ref{tab:ablation_relevance_diversity} compares the Relevance-only strategy with our diversity-aware strategy under different memory budgets.
These results demonstrate that query relevance alone is insufficient for frame packing.
By jointly considering relevance and visual diversity, \model allocates the limited high-resolution budget to more complementary frames, leading to more reliable performance under fixed memory constraints.

\paragraph{Ablation study on prompt relevance.}

We further analyze the effect of prompt relevance in the packing strategy on Qwen3.6-27B.
To isolate this factor, we construct a prompt-agnostic baseline,  \emph{Diversity-only}, which does not use the text prompt for frame selection. Instead, it uniformly samples $N$ frames from the video to preserve coarse temporal coverage. It then assigns frame priorities using a binary-tree midpoint ordering: the temporal midpoint of the sampled sequence is ranked first, followed by the midpoints of the left and right sub-intervals, and this process is applied recursively in a breadth-first manner.
Given this priority order, the same resolution allocation scheme is applied as in our method, with earlier-ranked frames receiving higher spatial resolutions.
Thus, Diversity-only preserves global temporal coverage but does not explicitly identify query-relevant evidence.

In contrast, our full strategy incorporates prompt relevance during frame sampling.
Table~\ref{tab:ablation_relevance_diversity} shows the comparison, where
our full strategy achieves stronger performance on most benchmarks and memory settings, especially on MLVU.
For example, under 512-frame memory, our strategy improves LongVideoBench accuracy from 74.0 to 74.9 with the same \texttt{d1f128d2f1536} scheme, and from 73.1 to 74.6 with \texttt{d1f384d2f512}.
These results show that temporal diversity alone is not sufficient: incorporating prompt relevance helps allocate high-resolution visual capacity to frames that are more likely to contain task-relevant evidence.

\begin{figure}[t]
\centering
\includegraphics[width=1.0\linewidth]{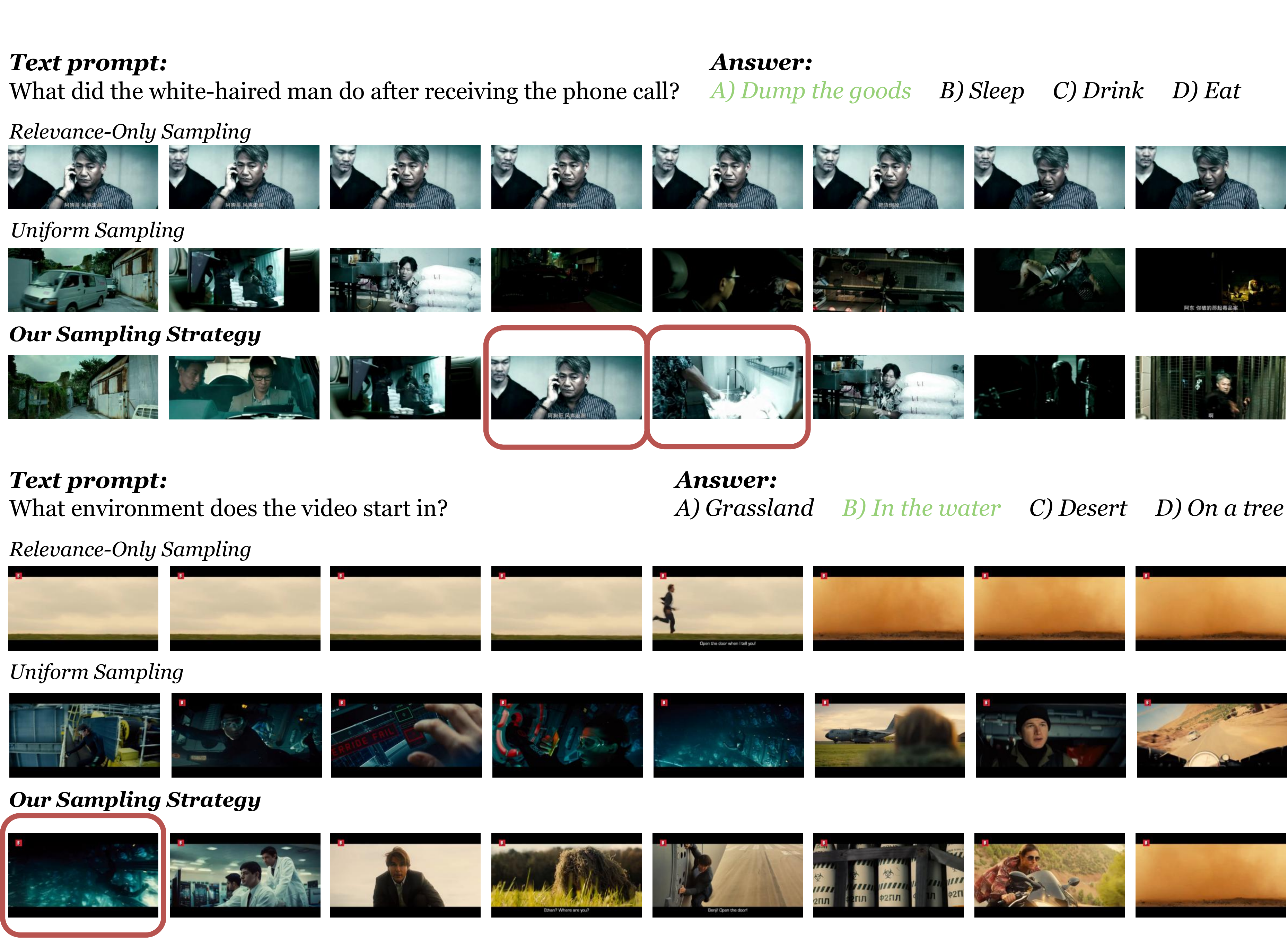}
\caption{\textbf{Qualitative comparison.} Relevance-only sampling selects frames based only on query-frame similarity and often produces redundant frames, while uniform sampling preserves temporal coverage but may miss question-relevant evidence. Our strategy combines prompt relevance and visual diversity, allowing it to select frames that better cover the video while retaining key evidence for answering the question. Red boxes highlight frames that contain important visual evidence.}
\label{fig:qualitative_results}
\end{figure}

\paragraph{Ablation study on model size.}

We further study how the effectiveness of \model changes with the scale of the base model.
To this end, we compare Qwen3.5-4B, Qwen3.5-9B, and Qwen3.6-27B.
For each model and memory budget, we select the best-performing configuration from the evaluated resolution schemes. The results are shown in Table~\ref{tab:ablation_model_size}.

Overall, the benefit of \model becomes more pronounced as the base model scales up.
On the smaller 4B and 9B models, \model can improve temporal coverage by packing more frames into the same memory budget, but the gains are less stable, especially under larger memory budgets.
In some cases (eg., Qwen3.5-4B on MLVU), increasing the number of packed frames does not lead to better performance and may even result in marginal degradation on certain benchmarks.
We attribute this to the capability boundary of smaller VLMs: although \model provides more diverse visual evidence, these models may not have sufficient capacity to effectively integrate and reason over a large number of visual observations.
This suggests that stronger VLMs are better able to exploit the additional temporal context and selectively attend to high-resolution, task-relevant evidence.
The advantage is especially evident when the memory budget is large, where \model can include more diverse contextual frames while still preserving fine-grained details for important frames.

\subsection{Qualitative results}
\label{subsec:qualitative}

\begin{figure}[t]
\centering
\includegraphics[width=1.0\linewidth]{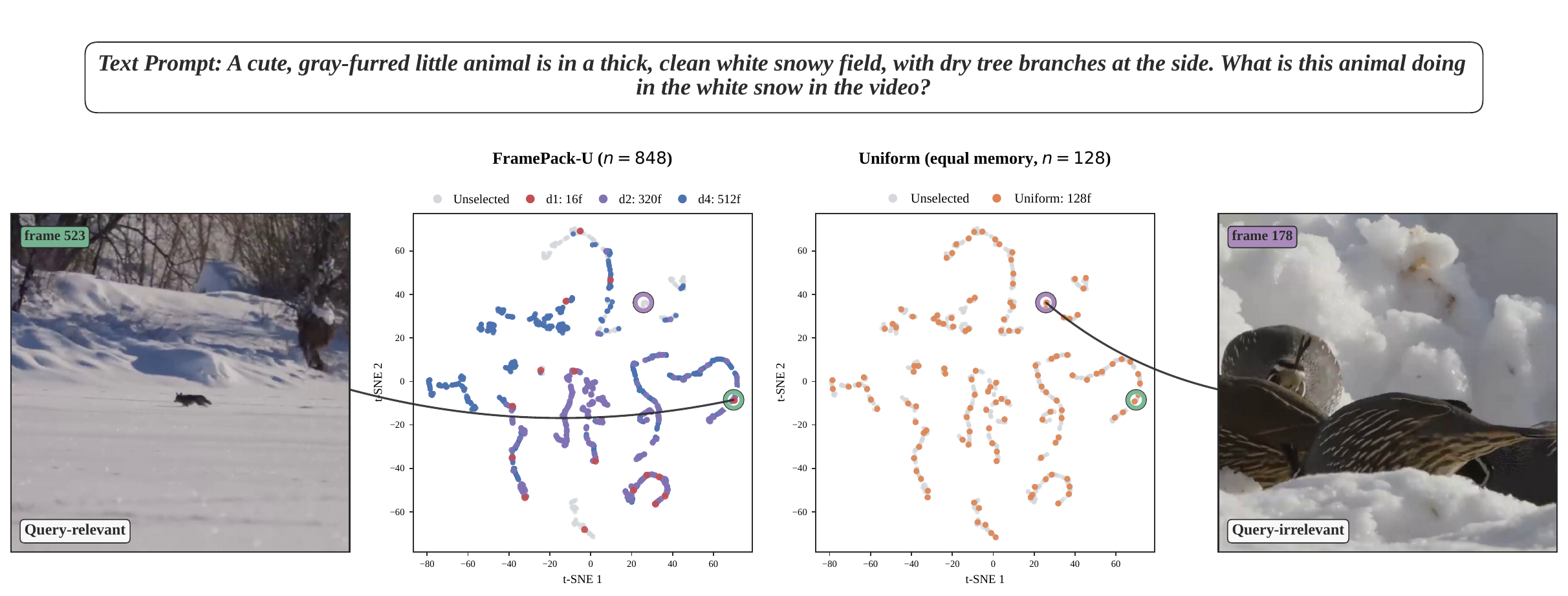}
% \vspace{-0.4em}
\includegraphics[width=1.0\linewidth]{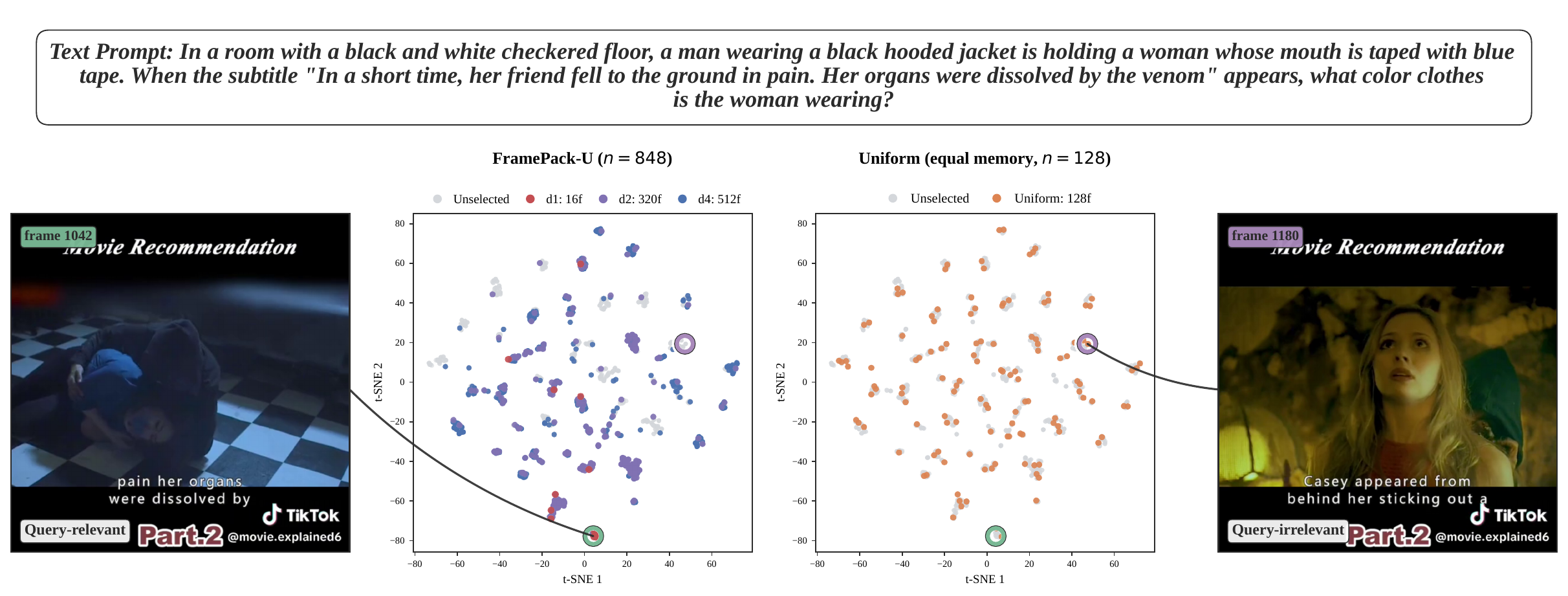}
% \vspace{-0.4em}
\includegraphics[width=1.0\linewidth]{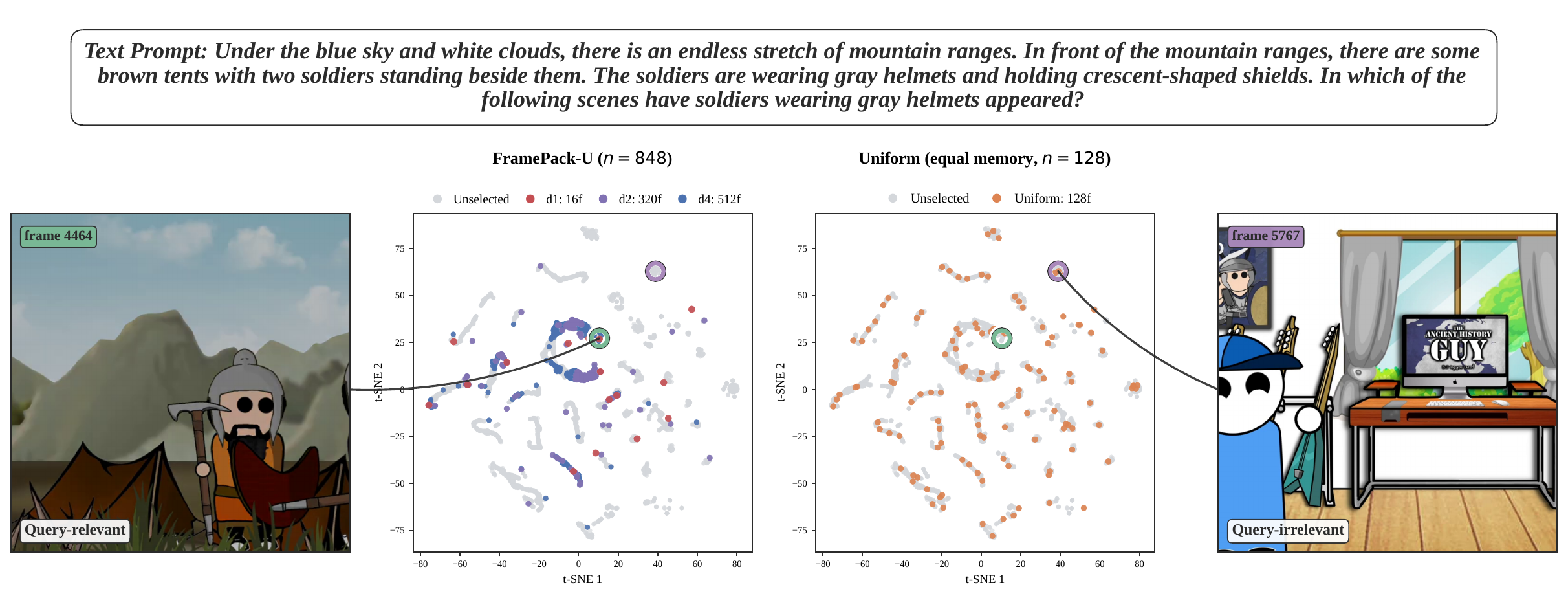}
% \vspace{-0.4em}
\caption{\textbf{t-SNE visualization.} 
We compare the feature distributions of frames selected by \model ($n=848$, scheme \texttt{d1f16d2f320d4f512}) against \textit{Uniform Sampling} ($n=128$, full resolution) across three distinct queries. Circles mark representative frames identified by each strategy, whose corresponding visual frames are displayed in the callout thumbnails. Points are colored according to their downsampling tiers ($d=1, 2, 4$).}
\label{fig:tsne}
\end{figure}

Fig~\ref{fig:qualitative_results} visualizes the sampled frames produced by different strategies.
Relevance-only sampling tends to repeatedly select frames that are highly similar to the text prompt, but these frames are often visually redundant and cover only a narrow temporal span.
Uniform sampling preserves coarse temporal coverage, but it is prompt-agnostic and may allocate visual budget to frames that are irrelevant to the question.
In contrast, our sampling strategy selects frames that are both instruction-relevant and visually diverse.
As shown in the examples, our method captures the key evidence needed to answer the question, such as the white-haired man dumping the goods after the phone call and the underwater scene at the beginning of the video.
These qualitative results further illustrate why combining prompt relevance with visual diversity leads to more effective frame packing for long video understanding.

To further understand how different sampling strategies distribute their selections over the visual content of a long video, we present t-SNE visualizations of frame visual features in Fig.~\ref{fig:tsne} under a fixed 128-frame memory budget. We compare \model with the standard \textit{Uniform Sampling} baseline ($n=128$). The visualization provides a feature-space view of how the two strategies differ in their frame selection patterns and in the semantic regions they cover.

Under uniform sampling, the selected frames are determined solely by their temporal positions. Since each frame is retained at full resolution ($d=1$), only $128$ frames can be accommodated under the given memory budget, resulting in a relatively sparse subset of the overall feature distribution. Consequently, some semantically distinctive or query-relevant regions may be under-represented, while the selected frames can correspond to content that is less useful for answering the question, as illustrated by the thumbnails on the right.

In contrast, by assigning multi-tier downsampling factors ($d=1,2,4$), \model packs $848$ frames within the same token memory footprint, yielding substantially denser coverage of the video's visual feature distribution. More importantly, its selections are guided by both query relevance and visual diversity rather than temporal position alone. As indicated by the cluster markers and the thumbnails on the left, \model tends to preserve frames from semantic regions that are more relevant to the text query, while the hierarchical resolution assignment further distinguishes their priorities: high-priority frames are retained at higher resolution ($d=1$), whereas lower-priority frames are compressed to $d=2$ or $d=4$ to provide broader contextual coverage. These visualizations therefore highlight two complementary properties of \model: it covers a broader range of the video's semantic content under the same memory budget, while also selecting and prioritizing frames differently from uniform sampling.

%% file: sections/latency.tex
% \section{Overhead and Latency Evaluations}
\section{Runtime analysis}
To clarify the preprocess overhead impact and study the end-to-end latency of our method, we conduct evaluations on LongVideoBench using Qwen3.5-4B, Qwen3.5-9B, and Qwen3.6-27B. To understand the scaling pattern as the number of frames increases, we report the average preprocess overhead of different video duration tiers and discuss the relativity of different inference stages. For both our method and vanilla models, we set fps=1.0 for candidate frames. We use the \texttt{d1f96d2f128} scheme for our method. All samples are evaluated on a single NVIDIA A800 GPU. Due to memory constraint, we set the maximum sampled frames of Qwen3.6-27B to 224 frames.

\begin{figure}[t]
\centering
\includegraphics[width=1.0\linewidth]{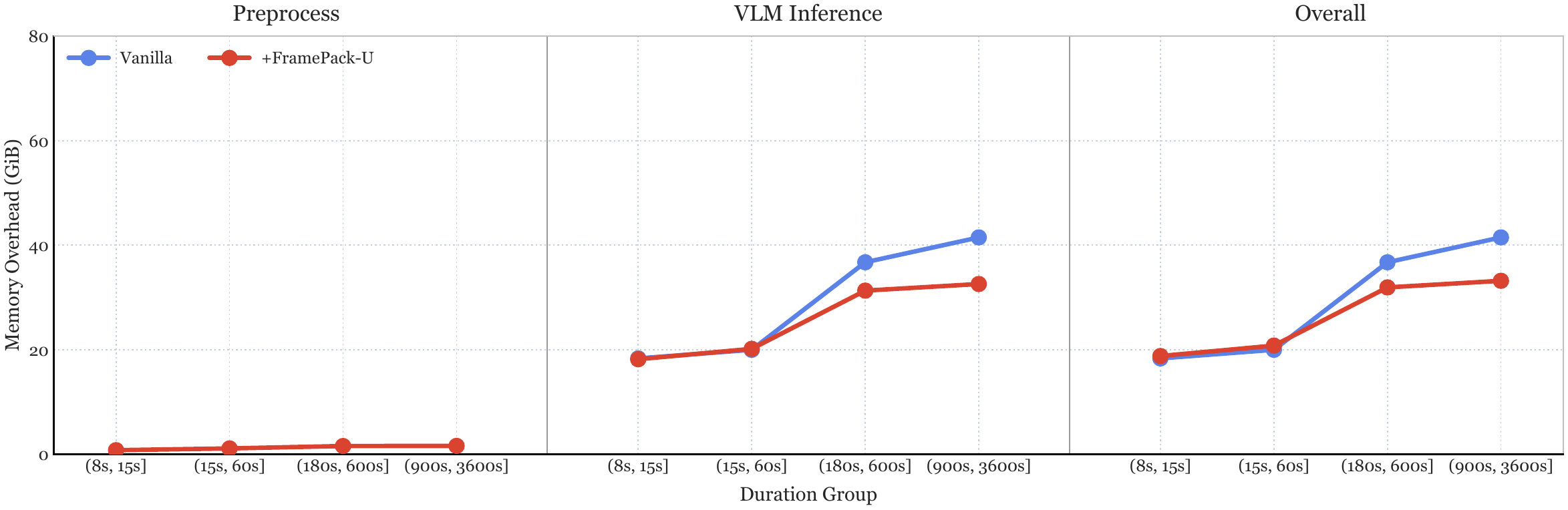}
\caption{\textbf{Memory overhead of \model on Qwen3.5-9B.}
We evaluate the preprocess, VLM inference and overall memory overhead of both vanilla models and \model on LongVideoBench. Results of different video duration tiers are reported. More results can be found at Table~\ref{tab:memory_overhead}.
% Noticeably, the vanilla Qwen3.6-27B model incurs Out-of-Memory error under fps=1 setting. Thus we sample a maximum of 224 frames, leading to a significantly lower memory overhead.
}
\label{fig:memory}
\end{figure}

\begin{figure}[t]
\centering
\includegraphics[width=1.0\linewidth]{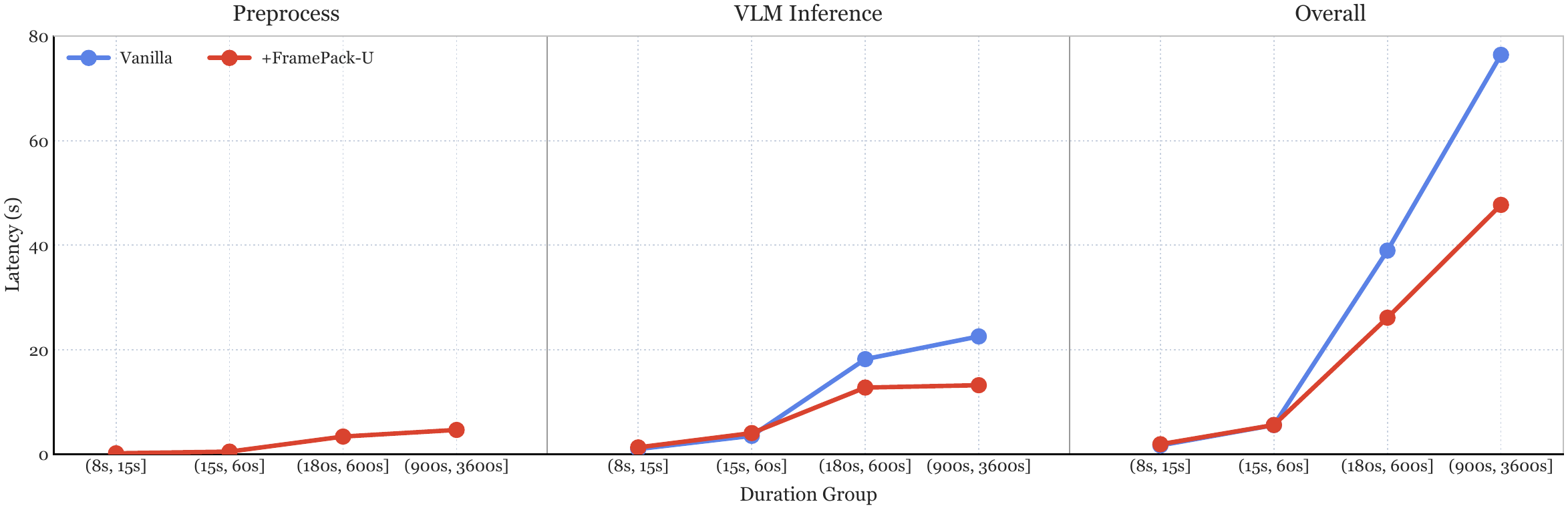}
\caption{\textbf{Latency evaluation of \model on Qwen3.5-9B.}
We evaluate the preprocess, VLM inference and overall latency of both vanilla models and \model on LongVideoBench. Results of different video duration groups are reported. More results can be found at Table~\ref{tab:inference_latency}.
% Noticeably, we restrict the vanilla Qwen3.6-27B model to sample a maximum of 224 frames for the memory constraint, leading to a significantly lower I/O latency.
}
\label{fig:latency}
\end{figure}

\begin{table*}[t]
\centering
% \scriptsize
% \small
\setlength{\tabcolsep}{4pt}
\renewcommand{\arraystretch}{1.08}
  \begin{tabular}{llccccc}
    \toprule
    \textbf{Stage} & \textbf{Model} & \textbf{15s} & \textbf{60s} & \textbf{600s} & \textbf{3600s} & \textbf{Avg.} \\
    \midrule
    % ----------------- Preprocess -----------------
    \multirow{6}{*}{Preprocess} 
      & \cellcolor{gray!15} Qwen3.5-4B                &\cellcolor{gray!15} --    & \cellcolor{gray!15}--    & \cellcolor{gray!15}--    &\cellcolor{gray!15} --    &\cellcolor{gray!15} -- \\
      & \quad + \model       & 0.779 & 1.117 & 1.576 & 1.602 & 1.288 \\
      & \cellcolor{gray!15}Qwen3.5-9B                & \cellcolor{gray!15}--    & \cellcolor{gray!15}--    & \cellcolor{gray!15}--    & \cellcolor{gray!15}--    & \cellcolor{gray!15}-- \\
      & \quad + \model       & 0.777 & 1.115 & 1.574 & 1.600 & 1.286 \\
      & \cellcolor{gray!15}Qwen3.6-27B               & \cellcolor{gray!15}--    & \cellcolor{gray!15}--    & \cellcolor{gray!15}--    & \cellcolor{gray!15}--    &\cellcolor{gray!15}-- \\
      & \quad + \model       & 0.775 & 1.115 & 1.575 & 1.600 & 1.307 \\
    \midrule
    % ----------------- VLM Inference -----------------
    \multirow{6}{*}{VLM Inference} 
      & \cellcolor{gray!15}Qwen3.5-4B                & \cellcolor{gray!15}9.207  & \cellcolor{gray!15}10.723 & \cellcolor{gray!15}26.174 & \cellcolor{gray!15}30.584 & \cellcolor{gray!15}19.860 \\
      & \quad + \model       & 9.033  & 10.871 & 21.145 & 22.311 & 16.294 \\
      \cmidrule(lr){2-7}
      & \cellcolor{gray!15}Qwen3.5-9B                & \cellcolor{gray!15}18.337 & \cellcolor{gray!15}19.980 & \cellcolor{gray!15}36.726 & \cellcolor{gray!15}41.483 & \cellcolor{gray!15}29.877 \\
      & \quad + \model       & 18.178 & 20.171 & 31.306 & 32.570 & 26.048 \\
      \cmidrule(lr){2-7}
      & \cellcolor{gray!15}Qwen3.6-27B               & \cellcolor{gray!15}52.217 & \cellcolor{gray!15}54.630 & \cellcolor{gray!15}73.137 & \cellcolor{gray!15}74.633 & \cellcolor{gray!15}64.473 \\
      & \quad + \model       & 52.139 & 55.065 & 71.420 & 73.275 & 63.697 \\
    \midrule
    % ----------------- Overall -----------------
    \multirow{6}{*}{Overall} 
      & \cellcolor{gray!15}Qwen3.5-4B                & \cellcolor{gray!15}9.207  & \cellcolor{gray!15}10.723 & \cellcolor{gray!15}26.174 & \cellcolor{gray!15}30.584 & \cellcolor{gray!15}19.860 \\
      & \quad + \model       & 9.650  & 11.488 & 21.762 & 22.928 & 16.911 \\
      \cmidrule(lr){2-7}
      & \cellcolor{gray!15}Qwen3.5-9B                & \cellcolor{gray!15}18.337 & \cellcolor{gray!15}19.980 & \cellcolor{gray!15}36.726 & \cellcolor{gray!15}41.483 & \cellcolor{gray!15}29.877 \\
      & \quad + \model       & 18.794 & 20.786 & 31.922 & 33.185 & 26.664 \\
      \cmidrule(lr){2-7}
      & \cellcolor{gray!15}Qwen3.6-27B               & \cellcolor{gray!15}52.217 & \cellcolor{gray!15}54.630 & \cellcolor{gray!15}73.137 & \cellcolor{gray!15}74.633 & \cellcolor{gray!15}64.473 \\
      & \quad + \model       & 52.755 & 55.680 & 72.035 & 73.890 & 64.313 \\
    \bottomrule
  \end{tabular}
  \caption{\textbf{Memory overhead (GiB) across different model sizes.} We evaluate the memory overhead of three inference stages on different video duration tiers of LongVideoBench. Preprocess stage represents the introduced CLIP-based frame ranking process. VLM Inference stage denotes the inference process of the off-the-shelf VLM. Overall results reflect the peak memory used during the end-to-end process.}
  \label{tab:memory_overhead}
\end{table*}

\begin{table*}[t]
\centering
% \scriptsize
% \small
\setlength{\tabcolsep}{4pt}
\renewcommand{\arraystretch}{1.08}
  \begin{tabular}{llccccc}
    \toprule
    \textbf{Stage} & \textbf{Model} & \textbf{15s} & \textbf{60s} & \textbf{600s} & \textbf{3600s} & \textbf{Avg.} \\
    \midrule
    % ----------------- Preprocess -----------------
    \multirow{6}{*}{Preprocess} 
      & \cellcolor{gray!15}Qwen3.5-4B                & \cellcolor{gray!15}--    & \cellcolor{gray!15}--    & \cellcolor{gray!15}--    & \cellcolor{gray!15}--    & \cellcolor{gray!15}-- \\
      & \quad + \model       & 0.183 & 0.518 & 3.460 & 4.729 & 2.354 \\
      & \cellcolor{gray!15}Qwen3.5-9B                & \cellcolor{gray!15}--    & \cellcolor{gray!15}--    & \cellcolor{gray!15}--    & \cellcolor{gray!15}--    & \cellcolor{gray!15}-- \\
      & \quad + \model       & 0.181 & 0.502 & 3.383 & 4.655 & 2.309 \\
      & \cellcolor{gray!15}Qwen3.6-27B               & \cellcolor{gray!15}--    & \cellcolor{gray!15}--    & \cellcolor{gray!15}--    & \cellcolor{gray!15}--    & \cellcolor{gray!15}-- \\
      & \quad + \model       & 0.169 & 0.480 & 3.210 & 4.366 & 2.178 \\
    \midrule
    % ----------------- VLM Inference -----------------
    \multirow{6}{*}{VLM Inference} 
      & \cellcolor{gray!15}Qwen3.5-4B                & \cellcolor{gray!15}0.787  & \cellcolor{gray!15}2.964  & \cellcolor{gray!15}14.322 & \cellcolor{gray!15}17.854 & \cellcolor{gray!15}9.486 \\
      & \quad + \model       & 1.027  & 3.370  & 9.878  & 9.933  & 6.336 \\
      \cmidrule(lr){2-7}
      & \cellcolor{gray!15}Qwen3.5-9B                & \cellcolor{gray!15}0.993  & \cellcolor{gray!15}3.523  & \cellcolor{gray!15}18.196 & \cellcolor{gray!15}22.543 & \cellcolor{gray!15}11.965 \\
      & \quad + \model       & 1.311  & 4.049  & 12.758 & 13.208 & 8.213 \\
      \cmidrule(lr){2-7}
      & \cellcolor{gray!15}Qwen3.6-27B               & \cellcolor{gray!15}1.857  & \cellcolor{gray!15}6.544  & \cellcolor{gray!15}33.778 & \cellcolor{gray!15}35.847 & \cellcolor{gray!15}20.710 \\
      & \quad + \model       & 2.335  & 7.503  & 31.914 & 34.030 & 20.022 \\
    \midrule
    % ----------------- Overall -----------------
    \multirow{6}{*}{Overall} 
      & \cellcolor{gray!15}Qwen3.5-4B                & \cellcolor{gray!15}1.490  & \cellcolor{gray!15}5.030  & \cellcolor{gray!15}35.143 & \cellcolor{gray!15}71.840 & \cellcolor{gray!15}29.744 \\
      & \quad + \model       & 1.639  & 4.971  & 23.501 & 45.093 & 19.639 \\
      \cmidrule(lr){2-7}
      & \cellcolor{gray!15}Qwen3.5-9B                & \cellcolor{gray!15}1.692  & \cellcolor{gray!15}5.584  & \cellcolor{gray!15}38.963 & \cellcolor{gray!15}76.421 & \cellcolor{gray!15}32.178 \\
      & \quad + \model       & 1.916  & 5.600  & 26.125 & 47.710 & 21.264 \\
      \cmidrule(lr){2-7}
      & \cellcolor{gray!15}Qwen3.6-27B               & \cellcolor{gray!15}2.541  & \cellcolor{gray!15}8.560  & \cellcolor{gray!15}48.875 & \cellcolor{gray!15}55.416 & \cellcolor{gray!15}30.364 \\
      & \quad + \model       & 2.917  & 8.986  & 44.652 & 66.767 & 32.402 \\
    \bottomrule
  \end{tabular}
  \caption{\textbf{Inference latency (s) across different model sizes.} We evaluate the memory overhead of three inference stages on different video duration tiers of LongVideoBench. Preprocess stage represents the introduced CLIP-based frame ranking process. VLM Inference stage denotes the inference process of the off-the-shelf VLM. Overall results reflect the total latency during the end-to-end process. Noticeably, we restrict the vanilla Qwen3.6-27B model to sample a maximum of 224 frames for the memory constraint, leading to a significantly lower I/O latency.}
  \label{tab:inference_latency}
\end{table*}

\paragraph{Memory overhead.} As shown in Table~\ref{tab:memory_overhead}, the memory overhead introduced by the CLIP-based scoring process is constant and relatively trivial. In comparison, the vanilla models suffer from a huge increase of the input tokens with frame scaling, while our method keeps a constant memory upper bound with even better performance.

\paragraph{Inference latency.} Table~\ref{tab:inference_latency} compares the inference latency of \model and the vanilla models. It can be observed that compared to the VLM inference stage, the latency of the CLIP-based scoring (preprocess) stage is relatively negligible. It is also noticeable that \model effectively reduces VLM inference and overall latency, making itself more efficient compared to the vanilla model. As for the results of Qwen3.6-27B, the higher overall latency for \model on extremely long videos stems directly from the I/O stage due to the difference in the number of loaded frames. Due to the GPU memory constraints of the massive Qwen3.6-27B model, the vanilla model is strictly restricted to sample a maximum of 224 frames. In contrast, \model evaluates candidate frames at fps=1.0, meaning it actually decodes up to 3600 frames for a 3600s video, while still safely operating within the constant memory budget.
Decoding 3600 frames from the disk incurs a significantly higher I/O overhead than decoding merely 224 frames. This massive hardware-bound I/O difference inevitably offsets the time saved during the VLM inference stage.
In a strictly equivalent comparison where both methods sample candidate frames at fps=1.0, \model would exhibit drastically lower overall latency, whereas the vanilla model would simply crash due to Out-of-Memory or suffer from catastrophic inference delays.

%% file: sections/limitation.tex
\section{Limitations}
\label{app:limitations}
Despite its simplicity and effectiveness, \model still has several limitations.
First, the benefit of frame packing depends on the capacity of the underlying VLM.
Although \model can pack more temporally diverse frames into the same memory budget, smaller models may not be able to effectively integrate and reason over the increased number of visual observations.
As discussed in Sec.~\ref{subsec:ablation}, this can lead to unstable gains under larger memory budgets.
For example, on smaller Qwen-series models, increasing the number of packed frames sometimes improves LongVideoBench but slightly degrades MLVU or Video-MME.
We attribute this failure case to the capability boundary of the base model rather than the packing strategy itself: when the model lacks sufficient visual reasoning capacity, providing more contextual frames can introduce additional distraction instead of useful evidence.

Second, \model is most naturally applicable to VLMs that encode video inputs as a sequence of independently processed image frames.
This design allows us to assign different spatial resolutions to different frames before visual encoding.
However, some recent video VLMs adopt grid-based or tightly coupled video encoders, where frames are first organized into a fixed spatiotemporal grid or processed jointly by a video-specific visual module.
For such models, e.g., Eagle2.5~\cite{eagle2_5}-style architectures, changing the resolution of individual frames may break the expected input format or require non-trivial modifications to the visual encoder and preprocessing pipeline.
As a result, \model is currently less plug-and-play for these grid-based video architectures.

Third, the current implementation relies on CLIP similarity as a lightweight proxy for prompt relevance.
While this choice keeps the method training-free and efficient, CLIP may fail to capture fine-grained temporal events, subtle interactions, or reasoning-heavy cues that are not easily recognized from individual frames.
In such cases, the relevance score may under-prioritize frames that are important for answering the question.
Similarly, the diversity term is based on visual feature similarity and may not fully reflect semantic complementarity.
Future work could explore stronger query-frame scoring functions or lightweight temporal-aware relevance models while preserving the training-free nature of the framework.

Finally, \model introduces a scheme-selection hyperparameter that controls the trade-off between temporal coverage and per-frame spatial fidelity.
Although our ablations show that the method is robust across a range of mixed-resolution schemes, the optimal allocation can still vary across models, benchmarks, and video domains.
A more adaptive strategy that automatically selects the packing scheme based on video length, query type, and model capacity would further improve usability and robustness.

%% file: sections/conclusion.tex
\section{Conclusion}
\label{sec:conclusion}

We introduced \model, a simple, training-free approach to long video understanding that achieves constant $O(1)$ memory complexity by packing frames into a fixed token budget at heterogeneous spatial resolutions. \model ranks frames by jointly considering instruction relevance and visual diversity, then allocates resolutions hierarchically so that the memory budget is concentrated on the most informative frames while full temporal coverage is preserved. Across LongVideoBench, MLVU, and Video-MME, \model consistently outperforms full-resolution baselines as well as prior token reduction and keyframe selection methods, and frequently matches or exceeds substantially larger proprietary VLMs. As a plug-and-play module for off-the-shelf VLMs, \model offers a practical step toward scalable long and streaming video understanding.

%% file: sections/appendix.tex
\appendix
In the appendix, we first provide detailed description of the alternative strategies in Sec.~\ref{app:alternative}. We also present additional analysis on \model in Sec.~\ref{app:more_ablation} and fine-grained evaluation results in Sec.~\ref{app:fg_results}. Finally, we discuss the limitations and failure cases in Sec.~\ref{app:limitations}.
\section{Alternative strategies}
\label{app:alternative}
\paragraph{Binary-tree midpoint ordering.}
In the absence of query information, frame importance can be estimated purely from the temporal structure of the video.  Inspired by the frame ordering of FramePack~\citep{framepack}, the most na\"ive approach assigns ranks in chronological order, giving full-resolution quota to the earliest frames. This is problematic because the first frames of a long video are not necessarily more informative than later
ones; in practice, this strategy reduces to uniform sampling of the early segment, discarding later temporal context entirely.

A more principled alternative is \emph{binary-tree midpoint ordering}. It assigns ranks recursively: the temporally central frame of the full video is ranked first, followed by the midpoints of the left and right sub-intervals, and so on, proceeding breadth-first across the resulting binary tree. More formally, let the frames be indexed as $\{1, \ldots, N\}$. The rank-$k$ frame is the midpoint of the sub-interval defined by the $k$-th node in a breadth-first traversal of the complete binary tree over the temporal axis, as detailed in Alg.~\ref{alg:bitree}. Such a design ensures that, for any budget $n_1 < N$, the $n_1$ anchor frames are distributed across the full temporal extent of the video rather than concentrated at one end, thus preserving global temporal coverage whilst allocating higher fidelity to temporally representative moments.

\begin{algorithm}[H]
\caption{Binary-Tree Midpoint Ordering}
\label{alg:bitree}
\begin{algorithmic}[1]
\Require Number of candidate frames $N$
\Ensure Query-agnostic priority index list $R=[R_1,R_2,\ldots,R_N]$

\State Initialize queue $Q \gets [(1,N)]$ \Comment{Each entry is a temporal interval $[l,r]$}
\State Initialize priority list $R \gets [\ ]$

\While{$Q$ is not empty}
    \State $(l,r) \gets Q.\mathrm{dequeue}()$
    \State $m \gets \lfloor (l+r)/2 \rfloor$ \Comment{Temporal midpoint}
    \State Append $m$ to $R$
    \If{$l \leq m-1$}
        \State $Q.\mathrm{enqueue}((l,m-1))$ \Comment{Left sub-interval}
    \EndIf
    \If{$m+1 \leq r$}
        \State $Q.\mathrm{enqueue}((m+1,r))$ \Comment{Right sub-interval}
    \EndIf
\EndWhile

\State \Return $R$
\end{algorithmic}
\end{algorithm}

\paragraph{CLIP-guided sampling.}
Another straightforward alternative is to rely only on text--frame relevance for frame sampling and ranking.
Given a text query $q$ and video frames $\{x_i\}_{i=1}^{T}$, we use CLIP~\citep{clip} to compute the cosine similarity between the text embedding of $q$ and the visual embedding of each frame:
\[
r_i = \cos\left(\phi_{\mathrm{img}}(x_i), \phi_{\mathrm{text}}(q)\right),
\]
where $\phi_{\mathrm{img}}(\cdot)$ and $\phi_{\mathrm{text}}(\cdot)$ denote the CLIP image and text encoders, respectively.
Frames are then sorted in descending order according to $r_i$, and the top $N$ frames are selected as candidates for packing.
The same ranking order is also used for resolution allocation: frames with higher CLIP similarity are assigned smaller downsampling factors and thus receive higher spatial resolutions, while lower-ranked frames are more aggressively compressed.

This strategy provides a simple query-aware baseline, as it explicitly prioritizes frames that are semantically aligned with the instruction.
However, since it only considers query relevance, it may repeatedly select visually similar frames when several adjacent frames all match the text query.
As a result, the selected high-resolution frames can become redundant and fail to cover complementary evidence across the video.
In contrast, our default strategy augments CLIP relevance with a diversity-aware greedy selection term, encouraging the packed frames to be both instruction-relevant and visually non-redundant.
\begin{algorithm}[H]
\caption{CLIP-Guided Sampling and Ranking}
\label{alg:clip_sampling}
\begin{algorithmic}[1]
\Require Video frames $\mathcal{V}=\{x_i\}_{i=1}^{T}$, text query $q$, number of packed frames $N$, packing scheme $\mathcal{S}=\{(d,f_d)\}_{d\in\{1,2,4,8,\ldots\}}$
\Ensure Packed visual sequence $\mathcal{V}^{*}$

\For{$i=1,\ldots,T$}
    \State Compute relevance score $r_i \gets \mathrm{CLIP}(x_i,q)$
\EndFor

\State Sort frame indices by descending relevance score:
\[
R \gets \mathrm{argsort}_{i\in\{1,\ldots,T\}}(r_i)[:N]
\]

\State Sort packing levels in $\mathcal{S}$ by increasing downsampling factor $d$
\State Initialize packed visual sequence $\mathcal{V}^{*}\gets [\ ]$

\For{$n=1,\ldots,N$}
    \State Find $d$ such that
    \(
    \sum_{d'<d} f_{d'} < n \leq \sum_{d'\leq d} f_{d'} .
    \)
    \State Pack frame $x_{R_n}$ as $\hat{x}_n \gets \mathrm{Resize}(x_{R_n}, 1/d)$
    \State Append $\hat{x}_n$ to $\mathcal{V}^{*}$
\EndFor

\State Reorder $\mathcal{V}^{*}$ according to the original temporal order of the corresponding frames
\State \Return $\mathcal{V}^{*}$
\end{algorithmic}
\end{algorithm}
\section{Additional analysis on \model}
\label{app:more_ablation}
\paragraph{Additional ablation study on schemes.}
\begin{table}[t]
\centering
\scriptsize
\setlength{\tabcolsep}{3pt}
\begin{tabular}{llcccc}
\toprule
Memory & Scheme & \#Frame & LongVideoBench & MLVU & Video-MME \\
\midrule
\rowcolor{gray!15}
\multirow{5}{*}{128 frames}
& Vanilla & 128 & 68.0 & 75.9 & 73.6 \\
& \texttt{d1f16d2f448} & 464 & 74.4 & 83.9 & 76.3 \\
& \texttt{d1f64d2f256} & 320 & 73.5 & 83.1 & 77.2 \\
& \texttt{d1f96d2f128} & 224 & 73.6 & 83.1 & \textbf{77.4} \\
& \texttt{d1f16d2f320d4f512} & 848 & \textbf{75.2} & \textbf{84.9} & 76.7 \\
\midrule
\rowcolor{gray!15}
\multirow{5}{*}{256 frames}
& Vanilla & 256 & 70.9 & 76.7 & 75.9 \\
& \texttt{d1f16d2f960} & 976 & 74.5 & \textbf{84.5} & 76.1 \\
& \texttt{d1f64d2f768} & 832 & \textbf{74.8} & 83.7 & 76.3 \\
& \texttt{d1f128d2f512} & 640 & 74.7 & 83.7 & 76.4 \\
& \texttt{d1f250d4f96} & 346 & 74.3 & 83.2 & \textbf{77.9} \\
\midrule
\rowcolor{gray!15}
\multirow{5}{*}{512 frames}
& Vanilla & 512 & 71.8 & 78.0 & 74.4 \\
& \texttt{d1f128d2f1536} & 1664 & \textbf{74.9} & 82.1 & 76.3 \\
& \texttt{d1f256d2f1024} & 1280 & 74.7 & 82.2 & 75.9 \\
& \texttt{d1f384d2f512} & 896 & 74.6 & \textbf{82.9} & 76.0 \\
& \texttt{d1f480d2f128} & 608 & 74.3 & 81.0 & \textbf{76.5} \\
\bottomrule
\end{tabular}
\caption{\textbf{Additional ablation study on resolution allocation schemes using Qwen3.6-27B.}
We report results under 128-, 256-, and 512-frame memory budgets, where each budget denotes the cost equivalent to processing the corresponding number of full-resolution frames. 
All mixed-resolution schemes use the diversity-aware sampling strategy. 
Bold numbers indicate the best result within each memory budget.}
\label{tab:full_ablation_schemes}
\end{table}
We further evaluate our method on various schemes across 128- and 256- frame memories. The full results are shown in Table \ref{tab:full_ablation_schemes}.

Under the 128-frame memory budget, all diversity-aware schemes bring clear improvements over the vanilla baseline.
Among them, \texttt{d1f16d2f320d4f512} achieves the best LongVideoBench and MLVU results, while \texttt{d1f96d2f128} obtains the best Video-MME score.
This suggests that increasing temporal coverage with more low-resolution frames is helpful, but allocating a small number of higher-resolution frames remains important for preserving fine-grained visual evidence.

A similar trend can be observed under the 256-frame memory budget.
Schemes such as \texttt{d1f16d2f960}, \texttt{d1f64d2f768}, and \texttt{d1f128d2f512} achieve comparable performance on LongVideoBench and MLVU, while d1f250d4f96 performs best on Video-MME.
This indicates that there is no single universally optimal allocation for all benchmarks; different tasks may prefer different balances between temporal coverage and spatial detail.

\paragraph{Additional ablation study on frame diversity.}
\begin{table}[t]
\centering
\scriptsize
\setlength{\tabcolsep}{3pt}
\begin{tabular}{llllll}
\toprule
Memory & Scheme & Strategy & LongVideoBench & MLVU & Video-MME \\
\midrule
\rowcolor{gray!15}
\multirow{10}{*}{128 frames}
& -- & Vanilla & 68.0 & 75.9 & 73.6 \\
& \multirow{3}{*}{\texttt{d1f16d2f448}}
& \model & \textbf{74.4} & \textbf{83.9} & 76.3 \\
& & \quad$-$ Relevance & 73.2 {\color{red}(-1.2)} & 81.5 {\color{red}(-2.4)} & 76.1 {\color{red}(-0.2)} \\
& & \quad$-$ Diversity & 72.8 {\color{red}(-1.6)} & 81.7 {\color{red}(-2.2)} & 68.9 {\color{red}(-7.4)} \\
\cmidrule(lr){3-6}
& \multirow{3}{*}{\texttt{d1f64d2f256}}
& \model & 73.5 & 83.1 & 77.2 \\
& & \quad$-$ Relevance & 72.2 {\color{red}(-1.3)} & 80.1 {\color{red}(-3.0)} & 76.2 {\color{red}(-1.0)} \\
& & \quad$-$ Diversity & 72.5 {\color{red}(-1.0)} & 81.0 {\color{red}(-2.1)} & 69.4 {\color{red}(-7.8)} \\
\cmidrule(lr){3-6}
& \multirow{3}{*}{\texttt{d1f96d2f128}}
& \model & 73.6 & 83.1 & \textbf{77.4} \\
& & \quad$-$ Relevance & 72.2 {\color{red}(-1.4)} & 78.3 {\color{red}(-4.8)} & 76.6 {\color{red}(-0.8)} \\
& & \quad$-$ Diversity & 72.1 {\color{red}(-1.5)} & 80.6 {\color{red}(-2.5)} & 68.8 {\color{red}(-8.6)} \\

\midrule
\rowcolor{gray!15}
\multirow{10}{*}{256 frames}
& -- & Vanilla & 70.9 & 76.7 & 75.9 \\
& \multirow{3}{*}{\texttt{d1f16d2f960}}
& \model & 74.5 & \textbf{84.5} & 76.1 \\
& & \quad$-$ Relevance & 74.5 {\color{gray}(+0.0)} & 82.0 {\color{red}(-2.5)} & \textbf{77.0} {\color{green!50!black}(+0.9)} \\
& & \quad$-$ Diversity & 73.6 {\color{red}(-0.9)} & 82.0 {\color{red}(-2.5)} & 69.7 {\color{red}(-6.4)} \\
\cmidrule(lr){3-6}
& \multirow{3}{*}{\texttt{d1f32d2f896}}
& \model & \textbf{74.8} & 83.7 & 76.0 \\
& & \quad$-$ Relevance & 74.6 {\color{red}(-0.2)} & 82.2 {\color{red}(-1.5)} & 76.7 {\color{green!50!black}(+0.7)} \\
& & \quad$-$ Diversity & 73.9 {\color{red}(-0.9)} & 81.8 {\color{red}(-1.9)} & 70.2 {\color{red}(-5.8)} \\
\cmidrule(lr){3-6}
& \multirow{3}{*}{\texttt{d1f128d2f512}}
& \model & 74.7 & 83.7 & 76.4 \\
& & \quad$-$ Relevance & 74.3 {\color{red}(-0.4)} & 81.9 {\color{red}(-1.8)} & 76.2 {\color{red}(-0.2)} \\
& & \quad$-$ Diversity & 73.7 {\color{red}(-1.0)} & 81.6 {\color{red}(-2.1)} & 70.0 {\color{red}(-6.4)} \\

\midrule
\rowcolor{gray!15}
\multirow{10}{*}{512 frames}
& -- & Vanilla & 71.8 & 78.0 & 74.4 \\
& \multirow{3}{*}{\texttt{d1f128d2f1536}}
& \model & \textbf{74.9} & 82.1 & \textbf{76.3} \\
& & \quad$-$ Relevance & 74.0 {\color{red}(-0.9)} & 82.3 {\color{green!50!black}(+0.2)} & \textbf{76.3} {\color{gray}(+0.0)} \\
& & \quad$-$ Diversity & 73.9 {\color{red}(-1.0)} & 80.4 {\color{red}(-1.7)} & 69.8 {\color{red}(-6.5)} \\
\cmidrule(lr){3-6}
& \multirow{3}{*}{\texttt{d1f256d2f1024}}
& \model & 74.7 & 82.2 & 75.9 \\
& & \quad$-$ Relevance & 74.6 {\color{red}(-0.1)} & 82.3 {\color{green!50!black}(+0.1)} & 76.0 {\color{green!50!black}(+0.1)} \\
& & \quad$-$ Diversity & 73.8 {\color{red}(-0.9)} & 80.6 {\color{red}(-1.6)} & 69.6 {\color{red}(-6.3)} \\
\cmidrule(lr){3-6}
& \multirow{3}{*}{\texttt{d1f384d2f512}}
& \model & 74.6 & \textbf{82.9} & 76.0 \\
& & \quad$-$ Relevance & 73.1 {\color{red}(-1.5)} & 81.1 {\color{red}(-1.8)} & 75.4 {\color{red}(-0.6)} \\
& & \quad$-$ Diversity & 73.7 {\color{red}(-0.9)} & 80.8 {\color{red}(-2.1)} & 69.9 {\color{red}(-6.1)} \\
\bottomrule
\end{tabular}
\caption{\textbf{Additional ablation study on prompt relevance and frame diversity using Qwen3.6-27B.} \textit{$-$ Relevance} removes prompt relevance and uses a prompt-agnostic diversity-based ordering, while \textit{$-$ Diversity} ranks frames only by query-frame similarity. \model combines prompt relevance and visual diversity. Values in parentheses denote the difference between each ablated variant and \model.}
\label{tab:full_ablation_relevance_diversity}
\end{table}

We provide additional ablation results on frame diversity under 128- and 256-frame memory budgets in Table~\ref{tab:full_ablation_relevance_diversity}.
Here, the \textit{- Diversity} variant removes the diversity-aware term and ranks frames only by query-frame similarity.
Although this variant remains query-aware, it tends to select visually similar frames when multiple adjacent frames are relevant to the prompt, causing redundant high-resolution allocation.

Under the 128-frame memory budget, removing diversity leads to clear degradation, especially on Video-MME.
For example, with \texttt{d1f16d2f448}, Video-MME drops from 76.3 to 68.9.
The same trend holds for \texttt{d1f64d2f256} and \texttt{d1f96d2f128}, where Video-MME decreases by 7.8 and 8.6 points, respectively.
Under the 256-frame memory budget, the effect remains consistent: removing diversity reduces Video-MME from 76.1 to 69.7 with \texttt{d1f16d2f960}, from 76.0 to 70.2 with \texttt{d1f32d2f896}, and from 76.4 to 70.0 with \texttt{d1f128d2f512}.
These results show that query relevance alone is insufficient for frame packing.
Visual diversity is critical for preventing redundant frames from occupying the limited high-resolution budget, thereby improving temporal coverage and robustness across benchmarks.

\paragraph{Additional ablation study on prompt relevance.}
We further analyze the effect of prompt relevance using the \textit{- Relevance} variant, which removes the text-query signal and uses a prompt-agnostic diversity-based ordering.
This baseline can preserve coarse temporal coverage, but it does not explicitly prioritize frames that contain task-relevant evidence.

Under the 128-frame memory budget, removing relevance consistently hurts LongVideoBench and MLVU.
For example, with \texttt{d1f16d2f448}, LongVideoBench decreases from 74.4 to 73.2 and MLVU decreases from 83.9 to 81.5.
With \texttt{d1f96d2f128}, the drop on MLVU becomes larger, from 83.1 to 78.3.
Under the 256-frame memory budget, \textit{- Relevance} occasionally matches or slightly exceeds \model on a single metric, such as Video-MME for \texttt{d1f16d2f960}, but it is less stable across benchmarks and usually performs worse on MLVU.
For instance, MLVU drops from 84.5 to 82.0 with \texttt{d1f16d2f960}, and from 83.7 to 82.2 with \texttt{d1f32d2f896}.
These results indicate that prompt relevance helps allocate high-resolution capacity to frames that are more likely to contain instruction-relevant evidence.
Combined with the diversity analysis above, this confirms that \model benefits from both components: prompt relevance identifies useful evidence, while frame diversity ensures non-redundant temporal coverage.

\paragraph{Additional ablation study on model size.}
\begin{table}[t]
\centering
\scriptsize
\setlength{\tabcolsep}{4pt}
\begin{tabular}{clccc}
\toprule
Memory & Model & LongVideoBench & MLVU & Video-MME \\
\midrule
\multirow{3}{*}{\makecell{128 frames \\ (\texttt{d1f96d2f128})}}
& Qwen3.5-4B & +1.7 & +4.0 & +1.1 \\
& Qwen3.5-9B & +2.8 & +3.0 & +0.6 \\
& Qwen3.6-27B & +5.6 & +7.2 & +3.8 \\
\midrule
\multirow{3}{*}{\makecell{256 frames \\ (\texttt{d1f64d2f768})}}

& Qwen3.5-4B & +0.2 & +1.6 & -0.9 \\
& Qwen3.5-9B & +1.1 & +1.4 & -0.2 \\
& Qwen3.6-27B & +3.9 & +7.0 & +0.4 \\
\midrule
\multirow{3}{*}{\makecell{512 frames \\ (\texttt{d1f256d2f1024})}}
& Qwen3.5-4B & +0.9 & -1.0 & -1.2 \\
& Qwen3.5-9B & +1.6 & -0.6 & -0.4 \\
& Qwen3.6-27B & +2.5 & +4.2 & +1.5 \\
\bottomrule
\end{tabular}
\caption{\textbf{Additional ablation study on model size.} We report the performance gain of \model over the vanilla baseline across different Qwen-series base models and memory budgets. Larger models generally obtain larger gains, suggesting that stronger VLMs can better exploit the packed multi-resolution visual evidence.}
\label{tab:full_ablation_model_size}
\end{table}

We provide additional results on different model sizes under 128-, 256-, and 512-frame memory budgets in Table~\ref{tab:full_ablation_model_size}.
We compare three Qwen-series backbones, including Qwen3.5-4B, Qwen3.5-9B, and Qwen3.6-27B, and report the performance gain of \model over the vanilla baseline under the same memory budget.

Overall, the benefit of \model becomes more pronounced as the base model scales up.
Under the 128-frame memory budget, all three models benefit from \model, but the gain is much larger on Qwen3.6-27B.
For example, Qwen3.6-27B improves by +5.6, +7.2, and +3.8 on LongVideoBench, MLVU, and Video-MME, respectively, while the gains on Qwen3.5-4B and Qwen3.5-9B are more moderate.
This suggests that stronger VLMs are better able to exploit the additional temporally diverse evidence introduced by frame packing.

The same trend is more evident under larger memory budgets.
At 256-frame memory, Qwen3.6-27B still obtains clear gains on LongVideoBench and MLVU, whereas smaller models show weaker or less stable improvements.
At 512-frame memory, Qwen3.6-27B consistently improves across all benchmarks, while Qwen3.5-4B and Qwen3.5-9B improve on LongVideoBench but show slight degradation on MLVU or Video-MME.
This indicates that when many more packed frames are provided, smaller VLMs may reach their capacity boundary and fail to effectively integrate the increased visual context.
Therefore, the occasional degradation on smaller models is likely due to the limited reasoning and visual integration capability of the base model, rather than the packing strategy itself.
These results further support that \model is especially beneficial for stronger VLMs, which can better leverage broad temporal coverage together with high-resolution evidence frames.

\section{Fine-grained results}
\label{app:fg_results}
\begin{table*}[t]
\centering
\scriptsize
\setlength{\tabcolsep}{4pt}
\renewcommand{\arraystretch}{1.08}
\begin{tabular}{lllcccc}
\toprule
Model & Setting & Method
& \makecell{LVB\\Overall}
& \makecell{LVB\\15 / 60 / 600 / 3600}
& \makecell{V-MME\\Overall}
& \makecell{V-MME\\S / M / L} \\
\midrule
\multirow{9}{*}{Qwen3.6-27B}
& \multirow{3}{*}{\makecell[l]{128-frame memory \\ \texttt{d1f16d2f448}}}
& \cellcolor{gray!15}Vanilla
& \cellcolor{gray!15}68.0
& \cellcolor{gray!15}82.0 / 83.1 / 68.9 / 58.0
& \cellcolor{gray!15}73.6
& \cellcolor{gray!15}85.2 / 74.1 / 61.4 \\
& & \model
& 74.4
& 82.5 / \textbf{84.9} / 73.5 / \textbf{69.1}
& 76.3
& 85.4 / 77.7 / 65.7 \\
& & $\Delta$
& {\color{green!50!black}+6.4}
& {\color{green!50!black}+0.5} / {\color{green!50!black}+1.8} / {\color{green!50!black}+4.6} / {\color{green!50!black}+11.1}
& {\color{green!50!black}+2.7}
& {\color{green!50!black}+0.2} / {\color{green!50!black}+3.6} / {\color{green!50!black}+4.3} \\
\cmidrule(lr){2-7}

& \multirow{3}{*}{\makecell[l]{256-frame memory \\ \texttt{d1f250d4f96}}}
& \cellcolor{gray!15}Vanilla
& \cellcolor{gray!15}70.9
& \cellcolor{gray!15}83.1 / 83.7 / 72.8 / 61.5
& \cellcolor{gray!15}75.9
& \cellcolor{gray!15}85.7 / 76.6 / 65.6 \\
& & \model
& 74.3
& 83.6 / 83.7 / 73.8 / 68.6
& \textbf{77.9}
& 86.1 / \textbf{79.2} / \textbf{68.3} \\
& & $\Delta$
& {\color{green!50!black}+3.4}
& {\color{green!50!black}+0.5} / {\color{gray}+0.0} / {\color{green!50!black}+1.0} / {\color{green!50!black}+7.1}
& {\color{green!50!black}+2.0}
& {\color{green!50!black}+0.4} / {\color{green!50!black}+2.6} / {\color{green!50!black}+2.7} \\
\cmidrule(lr){2-7}

& \multirow{3}{*}{\makecell[l]{512-frame memory \\ \texttt{d1f128d2f1536}}}
& \cellcolor{gray!15}Vanilla
& \cellcolor{gray!15}71.8
& \cellcolor{gray!15}82.0 / 83.1 / 72.3 / 64.5
& \cellcolor{gray!15}74.4
& \cellcolor{gray!15}\textbf{86.2} / 76.4 / 60.4 \\
& & \model
& \textbf{74.9}
& 83.1 / 82.6 / \textbf{76.5} / 68.8
& 76.3
& 84.7 / 77.4 / 66.7 \\
& & $\Delta$
& {\color{green!50!black}+3.1}
& {\color{green!50!black}+1.1} / {\color{red}-0.5} / {\color{green!50!black}+4.2} / {\color{green!50!black}+4.3}
& {\color{green!50!black}+1.9}
& {\color{red}-1.5} / {\color{green!50!black}+1.0} / {\color{green!50!black}+6.3} \\

\midrule
\multirow{9}{*}{Qwen3.5-35B-A3B}
& \multirow{3}{*}{\makecell[l]{128-frame memory \\ \texttt{d1f32d2f384}}}
& \cellcolor{gray!15}Vanilla
& \cellcolor{gray!15}59.4
& \cellcolor{gray!15}81.0 / 75.6 / 55.3 / 50.2
& \cellcolor{gray!15}66.7
& \cellcolor{gray!15}79.6 / 66.4 / 54.1 \\
& & \model
& 72.3
& 83.6 / 82.6 / 73.1 / 64.7
& 76.4
& 84.8 / 77.7 / 66.9 \\
& & $\Delta$
& {\color{green!50!black}+12.9}
& {\color{green!50!black}+2.6} / {\color{green!50!black}+7.0} / {\color{green!50!black}+17.8} / {\color{green!50!black}+14.5}
& {\color{green!50!black}+9.7}
& {\color{green!50!black}+5.2} / {\color{green!50!black}+11.3} / {\color{green!50!black}+12.8} \\
\cmidrule(lr){2-7}

& \multirow{3}{*}{\makecell[l]{256-frame memory \\ \texttt{d1f64d2f768}}}
& \cellcolor{gray!15}Vanilla
& \cellcolor{gray!15}60.3
& \cellcolor{gray!15}78.3 / 76.2 / 59.0 / 50.4
& \cellcolor{gray!15}67.1
& \cellcolor{gray!15}78.9 / 67.2 / 55.2 \\
& & \model
& 72.6
& 82.5 / 80.8 / 74.3 / 65.4
& 75.5
& 84.1 / 76.9 / 65.6 \\
& & $\Delta$
& {\color{green!50!black}+12.3}
& {\color{green!50!black}+4.2} / {\color{green!50!black}+4.6} / {\color{green!50!black}+15.3} / {\color{green!50!black}+15.0}
& {\color{green!50!black}+8.4}
& {\color{green!50!black}+5.2} / {\color{green!50!black}+9.7} / {\color{green!50!black}+10.4} \\
\cmidrule(lr){2-7}

& \multirow{3}{*}{\makecell[l]{512-frame memory \\ \texttt{d1f384d2f512}}}
& \cellcolor{gray!15}Vanilla
& \cellcolor{gray!15}59.8
& \cellcolor{gray!15}74.6 / 75.0 / 60.0 / 50.2
& \cellcolor{gray!15}67.3
& \cellcolor{gray!15}78.4 / 67.0 / 56.3 \\
& & \model
& 73.0
& \textbf{85.7} / 81.4 / 73.1 / 66.1
& 75.8
& 84.1 / 76.9 / 66.4 \\
& & $\Delta$
& {\color{green!50!black}+13.2}
& {\color{green!50!black}+11.1} / {\color{green!50!black}+6.4} / {\color{green!50!black}+13.1} / {\color{green!50!black}+15.9}
& {\color{green!50!black}+8.5}
& {\color{green!50!black}+5.7} / {\color{green!50!black}+9.9} / {\color{green!50!black}+10.1} \\
\bottomrule
\end{tabular}
\caption{\textbf{Fine-grained results on LongVideoBench and Video-MME.}
For LongVideoBench, we report the overall accuracy and the results on the 15s / 60s / 600s / 3600s subsets.
For Video-MME, we report the overall accuracy and the short / medium / long subsets.
Only the default diversity-aware packing strategy of \model is reported, with the vanilla baseline included as reference.
$\Delta$ denotes the difference between \model and Vanilla under the same setting.
Bold indicates the best result in each column or sub-column.}
\label{tab:fine_grained_results}
\end{table*}
We further report fine-grained results on the duration-based subsets of LongVideoBench and Video-MME.
For LongVideoBench, we report the overall accuracy and the results on the 15s, 60s, 600s, and 3600s subsets.
For Video-MME, we report the overall accuracy and the results on the short, medium, and long subsets.

As shown in Table~\ref{tab:fine_grained_results}, \model improves performance consistently across most fine-grained duration splits.
The gains are especially pronounced on longer videos, where preserving broader temporal context becomes more important.
On Qwen3.6-27B, under the 128-frame memory budget, \model improves LongVideoBench 3600s accuracy from 58.0 to 69.1, yielding a gain of +11.1.
It also improves the Video-MME long split from 61.4 to 65.7.
Under the 256-frame memory budget, the gains remain clear on long videos, with LongVideoBench 3600s improving from 61.5 to 68.6 and Video-MME long improving from 65.6 to 68.3.
These results indicate that multi-resolution packing is particularly helpful for long-duration understanding, where relevant evidence may appear sparsely across the video.

The improvement is even more significant on Qwen3.5-35B-A3B.
For example, under the 128-frame memory budget, \model improves LongVideoBench 600s and 3600s accuracy from 55.3 and 50.2 to 73.1 and 64.7, respectively.
It also improves Video-MME long from 54.1 to 66.9.
Under the 512-frame memory budget, \model still brings large gains on long-duration subsets, improving LongVideoBench 3600s from 50.2 to 66.1 and Video-MME long from 56.3 to 66.4.
These results suggest that \model effectively converts a fixed memory budget into broader temporal coverage, enabling the VLM to access more long-range evidence without increasing the visual memory cost.

\section{More comparisons with similar methods}
Token compression and key frame selection is a recognized paradigm in efficient video LLMs. Methods like SlowFast-LLaVA~\cite{xu2024slowfast}, Q-Frame~\cite{qframe} and LongVU~\cite{shen2024longvu} explore how to identify tokens or frames with key information in different ways. \model differs from these works in the following aspects:

\paragraph{SlowFast-LLaVA} relies on a rigid, query-agnostic two-stream architecture. Its resolution assignment is strictly dictated by a predetermined temporal frame rate. The Slow pathway operates at a low frame rate with high spatial resolution, while the Fast pathway utilizes a high frame rate with aggressive spatial pooling. It assigns resolution simply based on temporal position, completely ignoring whether the visual content is relevant to the instruction. In stark contrast, \model dynamically and deterministically assigns resolutions based on content relevance and visual diversity.
\paragraph{Q-Frame} assigns High, Medium, and Low resolutions using a probabilistic Gumbel-Max sampling trick applied to CLIP scores, followed by static hyperparameter thresholding to rank frames into resolution tiers. \model is mathematically governed by a strict constant memory constraint, formulated as Eq.~\ref{eq:budget}. Our resolution packing is not a heuristic ranking strategy. It is a deterministic, hierarchical optimization that mathematically guarantees O(1) memory complexity regardless of the video length.
\paragraph{LongVU} relies heavily on an extensive video-language fine-tuning stage, whereas \model operates entirely training-free on off-the-shelf VLMs. Methodologically, LongVU employs a disjointed pipeline that query-agnostically discards frames based on DINOv2~\cite{dinov2} visual similarity before introducing any cross-modal text queries. This creates a severe bottleneck that blindly destroys instruction-relevant evidence simply due to low-level visual redundancy. In contrast, \model's deterministic greedy algorithm evaluates semantic text-relevance and visual diversity simultaneously in a single, unified step. This joint optimization ensures that our strict memory budget is precisely allocated without prematurely dropping query-critical frames.

\paragraph{Empirical Comparison.} Empirically, we conduct controlled experiments to directly compare our strategy with Q-Frame and LongVU. For fairness, we adopt the same VLM (CLIP-ViT-B/32) for feature extraction and similarity calculation, and apply these methods on the same VLM (Qwen3.5-4B) under the same resolution scheme contract (\texttt{d1f96d2f128}). For the training-based method LongVU, we implement a LongVU-style frame ranking strategy for resolution allocation. These experiments control the same resolution assignment process and isolate the ranking strategy.

\begin{table*}[t]
\centering
\scriptsize
\setlength{\tabcolsep}{4pt}
\renewcommand{\arraystretch}{1.08}
  \begin{tabular}{lcccc}
    \toprule
    \textbf{Method} & \textbf{LVB} & \textbf{MLVU} & \textbf{VideoMME} & \textbf{Overall} \\
    \midrule
    % ---------------- Vanilla Model ----------------
    \cellcolor{gray!15} Vanilla Model & \cellcolor{gray!15}64.5 & \cellcolor{gray!15}70.3 & \cellcolor{gray!15}70.5 & \cellcolor{gray!15}68.4 \\
    \midrule
    % ---------------- Q-Frame ----------------
    \quad + Q-Frame & 62.6 & \textbf{76.4} & 66.1 & 68.4 \\
    \quad $\Delta$ & {\color{red}-1.9} & {\color{green!50!black}+6.1} & {\color{red}-4.4} & {\color{gray}+0.0} \\
    \cmidrule(lr){1-5}
    % ---------------- LongVU-Style Ranking ----------------
    \quad + LongVU-Style Ranking & 58.5 & 71.5 & 63.9 & 64.6 \\
    \quad $\Delta$ & {\color{red}-6.0} & {\color{green!50!black}+1.2} & {\color{red}-6.6} & {\color{red}-3.8} \\
    \cmidrule(lr){1-5}
    % ---------------- \model ----------------
    \quad + \model & \textbf{66.2} & 74.3 & \textbf{71.6} & \textbf{70.7} \\
    \quad $\Delta$ & {\color{green!50!black}+1.7} & {\color{green!50!black}+4.0} & {\color{green!50!black}+1.1} & {\color{green!50!black}+2.3} \\
    \bottomrule
  \end{tabular}
  \caption{\textbf{Performance comparison with similar token compression or key frame selection methods.} $\Delta$ denotes the performance gap with the vanilla model. Bold indicates the
  best result in each column.}
  \label{tab:more_comparison}
\end{table*}

Table~\ref{tab:more_comparison} shows that our method achieves superior performance, while other methods demonstrate a lack of robustness or even degradation. These results prove that our ranking strategy effectively distinguishes the most critical frames from less informative ones.